\documentclass[10pt,twocolumn,letterpaper]{article}

\usepackage{cvpr}
\usepackage{times}
\usepackage{epsfig}
\usepackage{graphicx}
\usepackage{amsmath}
\usepackage{amssymb}

\usepackage{authblk}
\newcommand*\samethanks[1][\value{footnote}]{\footnotemark[#1]}

\usepackage{tabularx}
\newcolumntype{L}[1]{>{\raggedright\arraybackslash}p{#1}}
\newcolumntype{C}[1]{>{\centering\arraybackslash}p{#1}}
\newcolumntype{R}[1]{>{\raggedleft\arraybackslash}p{#1}}

\usepackage{url}
\usepackage[caption = false]{subfig}
\usepackage{capt-of}

\usepackage[pagebackref=true,breaklinks=true,letterpaper=true,colorlinks,bookmarks=false]{hyperref}

\cvprfinalcopy 

\def\cvprPaperID{2287} 

\ifcvprfinal\fi
\begin{document}
\title{Controllable Descendant Face Synthesis}

\author{
Yong Zhang\textsuperscript{${\dag}$} \thanks{indicates equal contributions.${^\sharp}$indicates corresponding authors. This work was done when Le Li was an intern at Tencent AI Lab.} 
,~Le Li\textsuperscript{${\ddag}$${\dag}$}\samethanks
,~Zhilei Liu\textsuperscript{${\ddag}$${\sharp}$}
,~Baoyuan Wu\textsuperscript{${\dag}$} 
,~Yanbo Fan\textsuperscript{${\dag}$} 
,~Zhifeng Li\textsuperscript{${\dag}$}
\\
\textsuperscript{${\dag}$}Tencent AI Lab,
\textsuperscript{${\ddag}$}College of Intelligence and Computing, Tianjin University
\\
{\small \{zhangyong$201303$,wubaoyuan$1987$,fanyanbo$0124$\}@gmail.com,
\\
\small \{le\_li,zhileiliu\}@tju.edu.cn,michaelzfli@tencent.com}
}

\maketitle

%

\begin{abstract}
Kinship face synthesis is a topic raised to answer an interesting question that ``what will your future children look like?''
Very few works focus on this topic.
Existing methods model one-versus-one kin relation between only one parent face and one child face by directly using an auto-encoder without any explicit control over the resemblance of the synthesized face to the parent face.
Neither do they have control over age.
In this paper, we propose a novel method for controllable descendant face synthesis, which models two-versus-one kin relation between two parent faces and one child face.
Our model consists of an inheritance module and an attribute enhancement module.
The former is designed for accurate control over the resemblance between the synthesized face and parent faces.
The latter is designed for control over age and gender.
As there is no large scale database with father-mother-child kinship annotation, we propose an effective strategy to train the model without using the ground truth descendant faces.
No carefully designed image pairs are required for learning except only age and gender labels of training faces.
We evaluate the proposed method on three public benchmark databases to demonstrate its effectiveness through qualitative and quantitative evaluations.
\end{abstract}
\section{Introduction} \label{sec:intro}


Kinship between parents and children or between siblings has been studied for many years in the field of vision~\cite{debruine2009kin,dal2006kin} and psychology~\cite{kaminski2013children,alvergne2007differential}, which is valuable in applications such as finding missing children, criminal pursing, and social media analysis.
In the field of computer vision, most existing kinship-related works focus on kinship identification~\cite{georgopoulos2018modeling,dibeklioglu2017like,lu2017discriminative,lu2014neighborhood,yan2014discriminative}, \textit{i.e.,} justifying whether a given pair of faces has kinship. 
Very few works except  ~\cite{ertugrul2017will},~\cite{ertuugrul2018modeling}, and~\cite{ozkan2018kinshipgan} focus on kinship face synthesis, \textit{i.e.,} synthesizing a child face given a parent face, which is first proposed by Ertu{\u{g}}rul \textit{et al.}~\cite{ertugrul2017will}. 
Kinship face synthesis is referred as \textit{descendant face synthesis} in this paper since we focus on the kinship between parents and children. 

\begin{figure}
    \centering
    \includegraphics[width=0.95\linewidth]{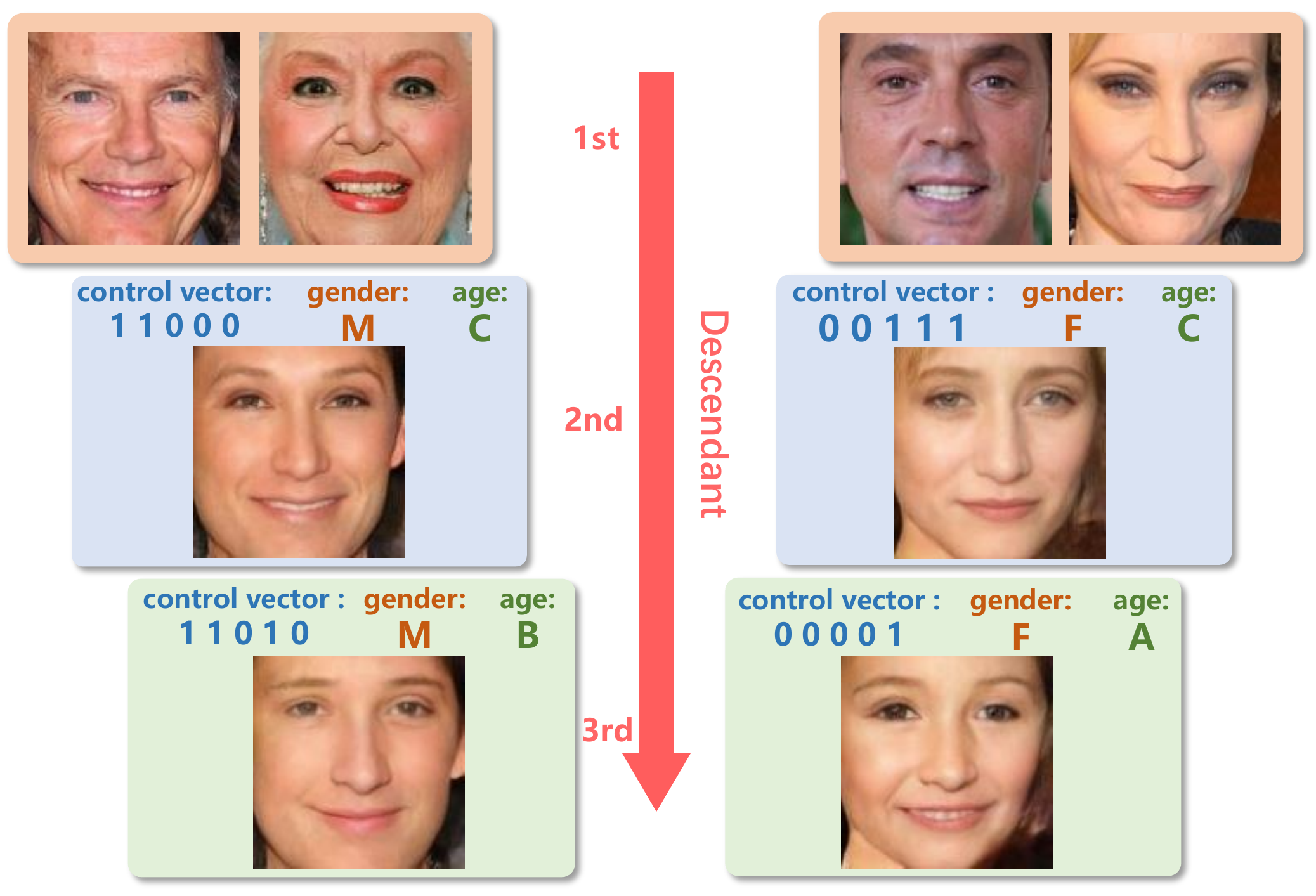}
    \caption{A tree of synthesized descendant faces. Faces of the 1st generation are given parent faces, while descendant faces of the 2nd and 3rd generations are synthesized by our method. The inheritance of facial components is determined by the control vector of which five bits correspond to left eye\&brow, right eye\&brow, nose, mouth, and profile, respectively. `0' means the component inherits from the male. `M' and `F' represent male and female. `A', `B', `C', and `D' represent
    age stages, respectively, \textit{i.e.,} 0-5, 6-15, 16-45, and larger than 45.   
    }
    \label{fig:teaser}
    \vspace{-5mm}
\end{figure}
\begin{figure*}[ht]
\centering
\subfloat[Existing framework of~\cite{ertuugrul2018modeling} and~\cite{ozkan2018kinshipgan}]{\label{fig:exising}\includegraphics[height=0.25\linewidth]{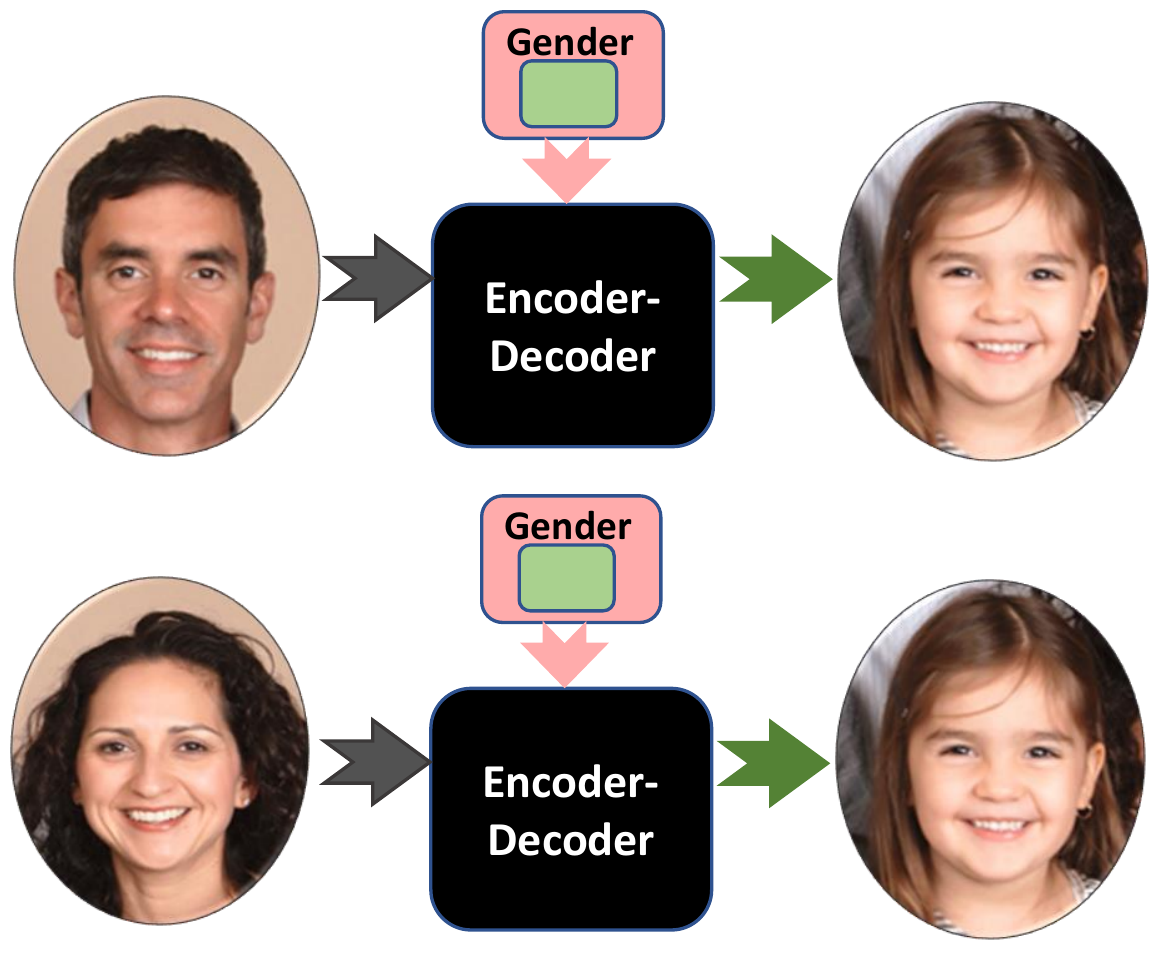}} 
\hspace{8mm}
\subfloat[Our framework]{\label{fig:ours}\includegraphics[height=0.25\linewidth]{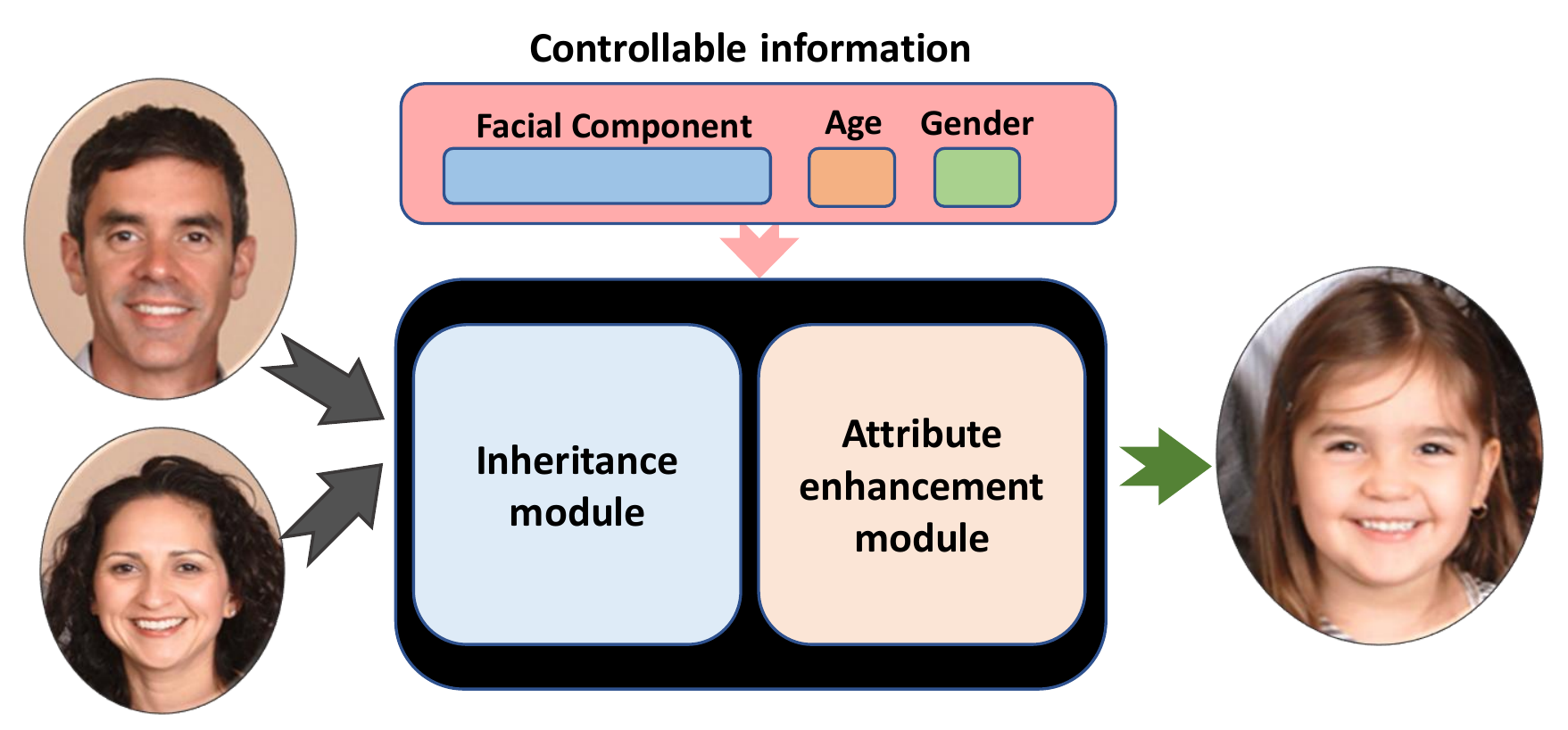}}

\vspace{1mm}
\caption{Comparison between our framework and the existing framework for descendant face synthesis. 
The existing framework of~\cite{ertuugrul2018modeling} and~\cite{ozkan2018kinshipgan} models the \textit{\textbf{one-versus-one}} relation (\textit{i.e.,} only one parent face and one child face) by directly using an auto-encoder or GAN.
Differently, our framework modes the \textit{\textbf{two-versus-one}} relation (\textit{i.e.,} two parent faces and one child face) by using two carefully designed modules. It has the exact control on the inheritance of facial components, age, and gender. 
}
\label{fig:methods}
\vspace{-5mm}
\end{figure*}

The drawbacks of these methods are summarized as follows. 
First, modeling one-versus-one relation ignores complementary information from the other parent face since child face has a resemblance to both parent faces~\cite{alvergne2007differential}. 
Second, they lack control over the resemblance of the synthesized face to parent faces because they implicitly learn the connection between one parent face and one child face without explicit emphasis on the resemblance.
Third, they have a fatal issue that in the training data one input face might correspond to multiple output faces because a couple could have several children under the same gender. This might mess up the model during training. 
Fourth, they have control over the gender of the synthesized face, but no control over the age.

To alleviate the above issues, we propose a novel method to model \textit{two-versus-one} relation between two parent faces and one child face for controllable descendant face synthesis based on generative adversarial networks. 
It has explicit control over the resemblance of facial components between the synthesized face and parent faces and also has control over age and gender. 
Note that the two-versus-one relation has been studied only for kinship verification~\cite{qin2015tri}, but has not been studied for descendant face synthesis. 
As shown in Fig.~\ref{fig:ours}, our framework consists of two modules, \textit{i.e.,} an inheritance module and an attribute enhancement module. 
The former is designed to control the resemblance of facial components between the synthesized face and parent faces. 
If a component of a child face resembles to that of the father face, it is referred as that the child inherits the component from the father. 
This module generates high-quality intermediate faces according to the control vector of the inheritance of facial components. 
Though a couple might have multiple children, the specification of inheritance almost makes a pair of parent faces correspond to only one child face during training, which alleviates the third issue above. 
The latter is designed for the enhancement of age and gender on the intermediate faces. 
Both modules are jointly learned in an end-to-end manner.

Currently, there is no large scale database with the kinship annotation of father-mother-child triplets. 
TSKinFace~\cite{qin2015tri} contains only 1015 tri-subject groups. 
Families in the Wild (FIW)~\cite{robinson2016families} has a large set of pairwise kinship annotations such as father-son, mother-son, etc., but it has only 2059 tri-subject groups.
They are not enough to train a deep net to model the two-versus-one relation. 
Hence, we propose an effective strategy for model learning without using the ground truth descendant faces by exploiting low-quality synthetic faces and the designed component exchange strategy.
Fig.~\ref{fig:teaser} shows the generated descendant faces of two generations with control over the inheritance of components, gender, and age by our method.

Our primary contributions are summarized as follows:
\begin{itemize}
	\item We propose a novel method to model two-versus-one kin relation for controllable descendant face synthesis. It has explicit control over the resemblance of facial components between the synthesized face and its parent faces and also has control over age and gender. 

    \item We propose an effective strategy for model learning by exploiting low-quality synthetic faces and the component exchange to compensate for the lack of a large scale database with father-mother-child kin annotation. 
	
\end{itemize}

\section{Related Work}


\noindent\textbf{Face synthesis.}
Great improvements have been achieved in several sub areas of face synthesis on basis of GANs~\cite{goodfellow2014generative}, including face reconstruction~\cite{huang2017beyond,lee2017unsupervised}, face swap~\cite{bitouk2008face,korshunova2016fast}, facial attribute manipulation~\cite{xiao2018elegant,Choi2018CVPR}, face makeup transfer~\cite{liu2016makeup,chang2018pairedcyclegan}, and face aging~\cite{zhang2017age,wang2018face}. 
These methods aim to modify local facial regions according to a specified attribute, to swap the whole face region, to transfer makeup from a specified template, or to generate faces at different age stages. 
However, they do not focus on generating descendant faces.  

\vspace{1mm}
\noindent\textbf{Kinship verification.}
Most previous studies on kin relation focus on kinship verification~\cite{georgopoulos2018modeling}, including pairwise kinship~\cite{wang2014leveraging,dehghan2014look,yan2014discriminative,ghahramani2014family,lu2014neighborhood,li2017kinnet,lu2017discriminative,BMVC2015_148,wang2018cross} and triplet-wise kinship~\cite{qin2015tri,fang2013kinship}. In order to use temporal information, video-based kinship verification methods are proposed in~\cite{dibeklioglu2013like} and~\cite{dibeklioglu2017like}. 
Kinship is also used to assist the learning of age progression in~\cite{shu2016kinship}. 
These methods aim to judge whether a given pair or triplet of faces has a kinship, rather than synthesizing a descendant face.

\vspace{1mm}
\noindent\textbf{Kinship (descendant) face synthesis.}
Very few works have studied kinship face synthesis except~\cite{ertugrul2017will},~\cite{ertuugrul2018modeling}, and~\cite{ozkan2018kinshipgan}.
~\cite{ozkan2018kinshipgan} uses a GAN for descendant face synthesis and uses a gender label to control the gender of the synthesized face. 
~\cite{ertuugrul2018modeling} uses four auto-encoders to model the relations of father-son, father-daughter, mother-son, and mother-daughter, respectively. 
The gender is controlled by the selection of one of the four auto-encoders. 
Both methods aim to generate one child face given only one parent face by modeling one-versus-one kin relation.  
The synthesized face is supposed to be the same as the ground truth child face. 
They simply treat the parent face as the input of an auto-encoder or GAN and use the child face as the output to learn a direct mapping between them as shown in Fig.~\ref{fig:exising}.
However, the image quality of visual results is poor in~\cite{ertugrul2017will},~\cite{ertuugrul2018modeling}, and~\cite{ozkan2018kinshipgan}. 
They do not perform well on keeping the resemblance between the parent face and the ground truth of the child face, which does not satisfy their original purpose. 
Compared with them, the main difference is that our method focuses on synthesizing a descendant face by modeling the two-versus-one relation with explicit control over the resemblance to the parent faces as well as control over age and gender, while~\cite{ertuugrul2018modeling} and~\cite{ozkan2018kinshipgan} model the one-versus-one relation by implicitly learning the mapping from one parent face to one child face without guarantee on the resemblance and the age.  

\section{The Proposed Approach}
\begin{figure*}[ht] 
    \centering
    \includegraphics[width=0.9\linewidth]{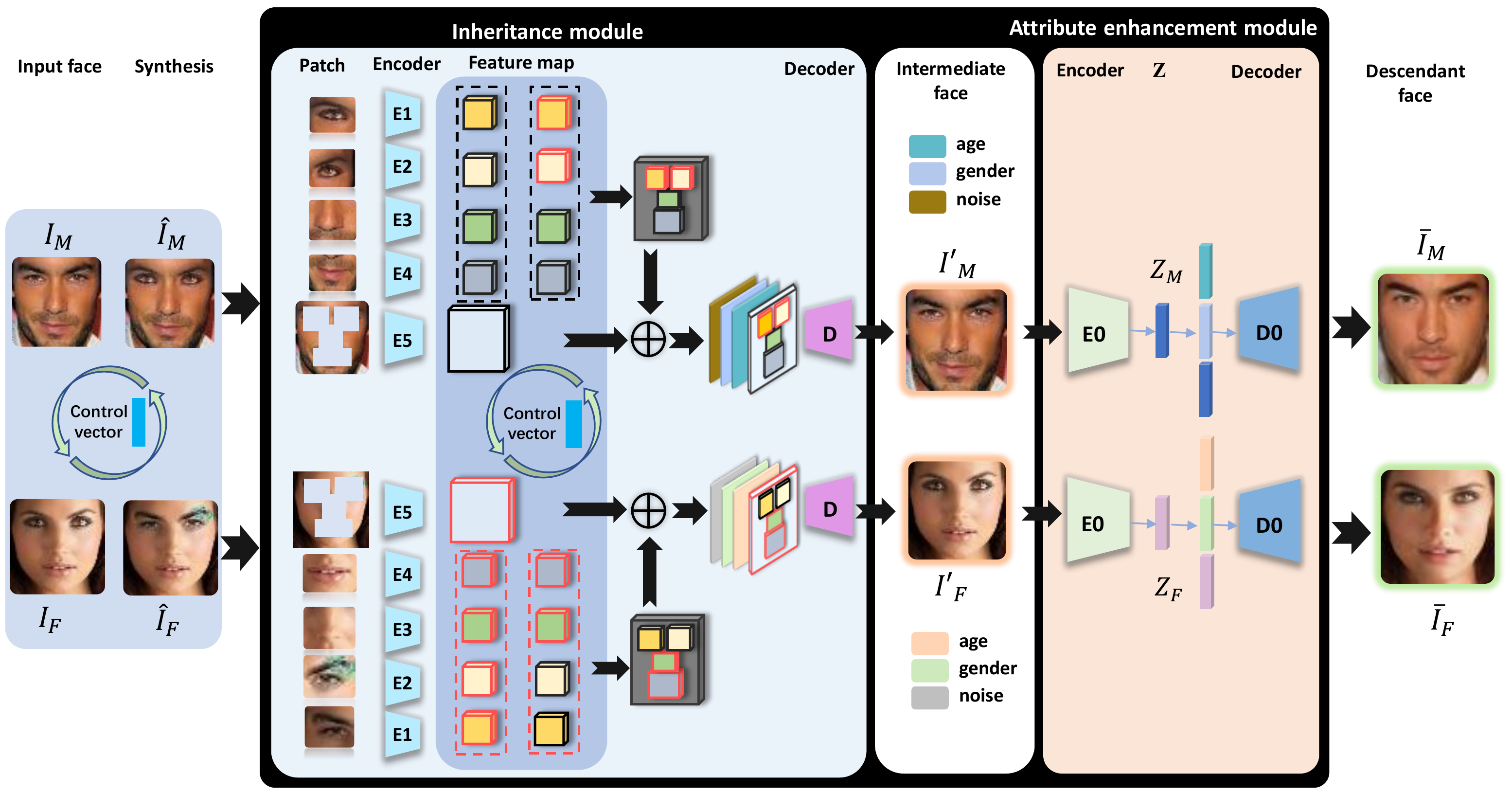}
    \caption{ The framework of the proposed method. The network consists of an inheritance module and an attribute enhancement module. \textit{ During training}, low quality synthetic faces are firstly generated according to the given parent faces and the control vector of inheritance. Then, they are fed into the inheritance module whose outputs are fed into the attribute enhancement module to generate final descendant faces. \textit{During testing}, the parent faces are directly fed into the inheritance module without generating synthetic faces. }
    \label{fig:proposed model}
    \vspace{-5mm}
\end{figure*}

We propose a novel method to model two-versus-one kin relation for descendant face synthesis with control over the resemblance of facial components between the synthesized face and its parent faces as well as control over age and gender. 
The framework of the proposed method is shown in Fig.~\ref{fig:proposed model}. 
We first introduce the strategy for learning without using the ground truth descendant faces in Sec.~\ref{subsection:Learning strategy}. 
Then, we present the structures of two modules in Sec.~\ref{subsection:Inheritance module} and Sec.~\ref{subsection:attribute enhancement module} followed by the designed losses in Sec.~\ref{subsection:Losses for joint learning of modules}.    

\vspace{-1mm}
\subsection{Learning without using the ground truth descendant faces} \label{subsection:Learning strategy}
\vspace{-1mm}
Previous studies mainly focus on kinship verification~\cite{georgopoulos2018modeling}, but very few works focus on descendant face synthesis except~\cite{ertuugrul2018modeling} and~\cite{ozkan2018kinshipgan}. 
Existing databases such as Families in the Wild (FIW)~\cite{robinson2016families}, TSKinFace~\cite{qin2015tri},  ~Sibling-Face~\cite{guo2014graph}, Family 101~\cite{fang2013kinship} and KinFaceW-I/II~\cite{lu2014neighborhood} are constructed for kinship verification. 
Since most works of kinship verification aim to identify pairwise kinship, most databases contain only pairwise kinship annotation.
They can be used for the one-versus-one descendant face synthesis~\cite{ertuugrul2018modeling,ozkan2018kinshipgan}, but are not applicable to our two-versus-one descendant face synthesis. 
The largest databases with the triplet-wise annotation of father-mother-child are FIW and TSKinFace which contain only 2059 and 1015 tri-subject groups, respectively. 
They are not enough to train a deep model that contains millions of parameters. 

We propose a strategy for learning without the ground truth of descendant faces by decomposing the task into two sub tasks and leveraging low-quality synthetic faces.
One is to take control over the resemblance of facial components between the synthesized face and its parent faces, 
\textit{i.e.,} the inheritance module. 
The other is to take control over age and gender, \textit{i.e.,} the attribute enhancement module. 
To supervise the learning of the inheritance module, we exchange facial components of parent faces according to the control vector of inheritance to generate synthetic faces.
The selected components of one parent face are replaced with the corresponding components from the other parent face by using color correlation~\cite{colorcorrection}. 
Each patch of each component is divided by a Gaussian blur of itself and then multiplied by a Gaussian blur of the target face.  
Note that the quality of synthetic faces is low since there are noticeable artifacts around facial components. 
We use the low quality synthetic faces as the input of the inheritance module. 

Inside the inheritance module, facial components will be exchanged back according to the control vector in the latent space. 
The intermediate face generated by the decoder will be compared to the original face to provide supervision.
Let ${I}_M$ denote the input male parent face and ${I}_F$ denote the female one. 
Let $\hat{I}_M$ and $\hat{I}_F$ denote faces after component exchange by color correlation~\cite{colorcorrection}. 
$\mathbf{v}$ is a 5-bit binary control vector of the inheritance, of which bits correspond to facial components, including left eye\&brow, right eye\&brow, nose, mouth and profile. 
$\mathbf{v}_i = 0$ ($i\in {1,2,3,4,5}$) means the $i$-th facial component inherits from the male face while $\mathbf{v}_i = 1$ means it inherits from the female face. 
`eye\&brow' means eye and brow are included in one patch.
Let $y^M_a$ and $y^M_g$ denote the age and gender of the male face and $y^F_a$ and $y^F_g$ for the female. 
The generation of synthetic faces by component exchange can be represented as 
\begin{align}
    \hat{I}_M, \hat{I}_F &= f_{\text{syn}}(I_M, I_F,\mathbf{v},y^M_a, y^M_g, y^F_a, y^F_g), 
\end{align}
where $\hat{I}_M$ and $\hat{I}_F$ are the inputs of the inheritance module. 

{Please note that if a large scale database with the father-mother-child kinship annotation is available, our method can be easily extended to exploit the ground truth descendant faces by adding a reconstruction loss between them and the generated descendant faces instead of using the low-quality synthetic faces as input.}

\subsection{Inheritance module} \label{subsection:Inheritance module}
The inheritance module is designed to control the resemblance of facial components between the synthesized face and its parent faces.
The inputs of the module consist of three parts, \textit{i.e.,} a pair of parent faces, the control vector of inheritance, and the age and gender of each parent face. 
As shown in Fig.~\ref{fig:proposed model}, parent faces are firstly decomposed into five facial components according to facial landmarks. 
Each component is represented by a patch. 
Note that the profile is the face image whose components are filled with black masks.
Then each component is fed into an encoder individually to get its feature map. 
Since components have different appearances, we use individual encoders to capture their specific features of shape, color, and texture.

The inheritance of facial components is performed by the exchange of feature maps between the female and the male according to the control vector in the latent space. 
Two combinations of feature maps can be generated through the component exchange.
One follows the control vector and the other follows the inverse of the control vector.
The feature maps of each combination are integrated into a new feature map according to their positions in the input face.
To incorporate the information of age and gender, we expand the labels of age and gender to two feature maps as the same size as the integrated feature map.  
Then these feature maps are concatenated with a noise feature map and fed into a decoder to generate an intermediate face.  
Let $I_M'$ denote the output male face of the inheritance module and $I_F'$ denote the female.  
The inheritance module can be represented as 
\begin{align}
       I_M', I_F' &= f_{\text{inh}}(\hat{I}_M, \hat{I}_F, \mathbf{v}, y^M_a, y^M_g, y^F_a, y^F_g). 
\end{align}

\subsection{Attribute enhancement module} \label{subsection:attribute enhancement module}

The attribute enhancement module is used to enhance gender and age on the intermediate faces from the inheritance module. 
The intermediate faces are encoded into a latent space by an encoder. 
The latent features are concatenated with the expanded vectors of age and gender and then fed into a decoder to generate the final descendant faces. 
The attribute enhancement module can be represented as 
\begin{align}
    \bar{I}_M, \bar{I}_F &= f_{\text{att}}(I_M', I_F', y^M_a, y^M_g, y^F_a, y^F_g), 
\end{align}
where $\bar{I}_M$ and $\bar{I}_F$ are the final descendant faces.

\subsection{Losses for joint learning of both modules} \label{subsection:Losses for joint learning of modules}

WGAN~\cite{Gulrajani2017} is used as the generator of the inheritance module for its stability of training. 
The adversarial loss is 
\begin{align}
   L_{\text{inh}}^{\text{adv}} &= \sum_{s \in \{M,F\}}
   \mathbb{E}_{{I}'_{s}\sim p({I}'_{s})} \left [ D_{I} \left ( {I}'_{s} \right ) \right] 
   - \mathbb{E}_{I_{s}\sim p_{data}({I_{s}})} \left [ D_{I} \left ( I_{s} \right ) \right ] \nonumber\\
    & ~~~~~~ + \lambda_{gp} \mathbb{E}_{\tilde{I}_{s}\sim p(\tilde{I}_{s})}\left [ \left ( \left \|\bigtriangledown _{\tilde{I}_{s}} D_{I}\left ( \tilde{I}_{s} \right ) \right \|_2-1 \right )^2 \right ] ,  \nonumber
\end{align} 
where $\tilde{I}_g$ is a randomly sampled image and 
$\lambda_{gp}$ is the hyperparameter of WGAN. 
$D_{I}$ is the discriminator to distinguish real images from fake ones. 
As in~\cite{ShrivastavaPTSW16}, it outputs a 2 $\times$ 2 probability map instead of a single scalar value.
As shown in Sec.~\ref{subsection:Learning strategy}, by using synthetic faces and the exchange of components, the difference between the output faces of the inheritance module and the original faces can be used to provide supervision. 
The pixel-wise loss is defined as 
\begin{equation}
L_{\text{inh}}^{\text{pix}} = \sum_{s\in \{M,F\}} \mathbb{E}_{I'_s , I_s} \left [ \left\|I'_s-I_s\right\|_2 \right],   \label{equation: pixel_loss}
\end{equation}
where $I'_s \sim p(I'_s)$ and $I_s \sim p_{\text{data}}(I_s)$. 
Since facial components of the intermediate face inherit from parent faces which could be very different in appearance and age, we use the information of age and gender to improve their consistency. 
We use ResNet18~\cite{he2016deep} to build a pre-trained age classifier and a gender classifier to constrain the generator. 
The losses are defined as 
\begin{equation}
L_{\text{inh}}^{\text{age}}= \sum_{s\in\{M,F\}}\mathbb{E}_{{I}'_{s}\sim p({I}'_{s})} \left [ \left\|D_{a} \left ( {I}'_{s} \right )-y_a^{s}\right\|_2 \right ],
\end{equation}
\begin{equation}
L_{\text{inh}}^{\text{gen}} = \sum_{s\in\{M,F\}}\mathbb{E}_{{I}'_{s}\sim p({I}'_{s})} \left [ \left\|D_g \left ( {I}'_{s} \right )-y_g^{s}\right\|_2 \right ],
\end{equation}
where $D_g$ is the gender classifier and $D_a$ is the age classifier to classify four age stages (`infant’, `teen’,
`adult’, and `older adult’) , \textit{i.e.,} `A' (0-5), `B' (6-15), `C' (16-45), `D' ($>45$). 
Note that the age can be divided into more groups with only a minor change in the number of output neurons of the age classifier.  
$y_a^s$ and $y_g^s$ are the age and gender labels of the input face $I_s$. 
Besides, we use the pre-trained 19-layer VGG to compute perceptual loss~\cite{simonyan2014very} to gain more facial details. 
The perceptual loss is defined as 
\begin{align} 
L_{\text{inh}}^{\text{per}}=& \sum_{s\in\{M,F\}} \mathbb{E}_{I'_s , I_s} \left [
\left\|f^{2,2}_{I_{s}}-f^{2,2}_{{I}'_s}\right\|_2 + 
\left\|f^{5,4}_{I_{s}}-f^{5,4}_{{I}'_s}\right\|_2 \right ],\nonumber
\end{align}
where $f_{I_s}^{i,j}$ is the feature map obtained by the $j$-th convolution layer before the $i$-th maxpooling layer in VGG19.  
The total loss of the inheritance module is computed as 
\begin{align}
L_{\text{inh}} &=  L_{\text{inh}}^{\text{age}} + L_{\text{inh}}^{\text{gen}} + \lambda_{11} L_{\text{inh}}^{\text{pix}} + \lambda_{12} L_{\text{inh}}^{\text{adv}} + 
\lambda_{13} L_{\text{inh}}^{\text{per}}.
\end{align}

The conditional auto-encoder~\cite{baldi2012autoencoders} is augmented with a discriminator to distinguish real images from fake ones in CAAE~\cite{zhang2017age}. 
However, the generated images are generally ambiguous. 
Inspired by~\cite{zhang2017age}, to better preserve the identity and improve the quality of generated faces, we enhance CAAE with a perceptual loss for modeling both age and gender in the attribute enhancement module.
The loss of the discriminator is defined as 
\begin{scriptsize}
\begin{align} \small
    L_{\text{att}}^{\text{adv}}= \sum_{s \in \{M,F\}} \mathbb{E}_{\bar{I}_s \sim p(\bar{I}_s)} \left[ \log D_{\bar{I}} \left ( \bar{I}_{s}\right ) \right ] + \mathbb{E}_{{I}_s \sim p({I}_s)} \left[ \log \left ( 1 - D_{\bar{I}} \left ( {I}_{s}\right ) \right ) \right ], \nonumber
\end{align}
\end{scriptsize}
where $D_{\bar{I}}$ is the discriminator to discriminate real faces from synthesized descendant faces. 
The reconstruction loss is defined by using the pixel-wise difference between the synthesized faces and the input faces, \textit{i.e.,} 
\begin{equation}
    \begin{split}
        L_{\text{att}}^{\text{pix}}=& \sum_{s\in \{M,F\}} \mathbb{E}_{I'_s , \bar{I}_s} \left[ \left\|\bar{I}_{s}-{I}'_s\right\|_2 \right]. 
    \end{split}
\end{equation}
As the same as the definition of $L_{\text{inh}}^{\text{per}}$, the perceptual loss is
\begin{align}
L_{\text{att}}^{\text{per}}=& \sum_{s\in {M,F}} \mathbb{E}_{I'_s , \bar{I}_s} \left[ \left\|f^{2,2}_{\bar{I}_{s}}-f^{2,2}_{{I}'_s}\right\|_2 +
\left\|f^{5,4}_{\bar{I}_{s}}-f^{5,4}_{{I}'_s}\right\|_2 \right]. \nonumber
\end{align}
The total loss of the attribute enhancement module is  
\begin{align}
L_{\text{att}} = L_{\text{att}}^{\text{pix}} + \lambda_{21} L_{\text{att}}^{\text{adv}} + 
\lambda_{22} L_{\text{att}}^{\text{per}}.
\end{align}

The full objective function of the joint learning of both modules is defined as 
\begin{align}
L = L_{\text{inh}} + \lambda_0 L_{\text{att}}. 
\end{align}

\section{Experiments}
\subsection{Settings}
\noindent \textbf{Datasets.}
CelebAHQ~\cite{karras2017progressive} is a high-resolution database from which 11,915 female subjects  and 7,756 male subjects are collected for our task. 
SiblingDB~\cite{vieira2014detecting} is a high-resolution database from which 77 female subjects and 103 male subjects are collected.
Each subject has one image.
We use $90\%$ of images in both databases for training and the $10\%$ left for testing. 
There is no overlap between the training and testing sets. 
TSKinFace ~\cite{qin2015tri} is a database with the annotation father-mother-child kin relationship, which only contains 1015 tri-subject groups. 
It is used to compare the generated faces with the ground truth children faces. 
Note that our method does not require the input pair of parents to be a true couple during training. 
So the face of any male and the face of any female can be treated as a pair to feed into our network, which enables us to construct a large set of pairs for model learning.

\begin{figure*}
    \centering
    \includegraphics[width=1\linewidth]{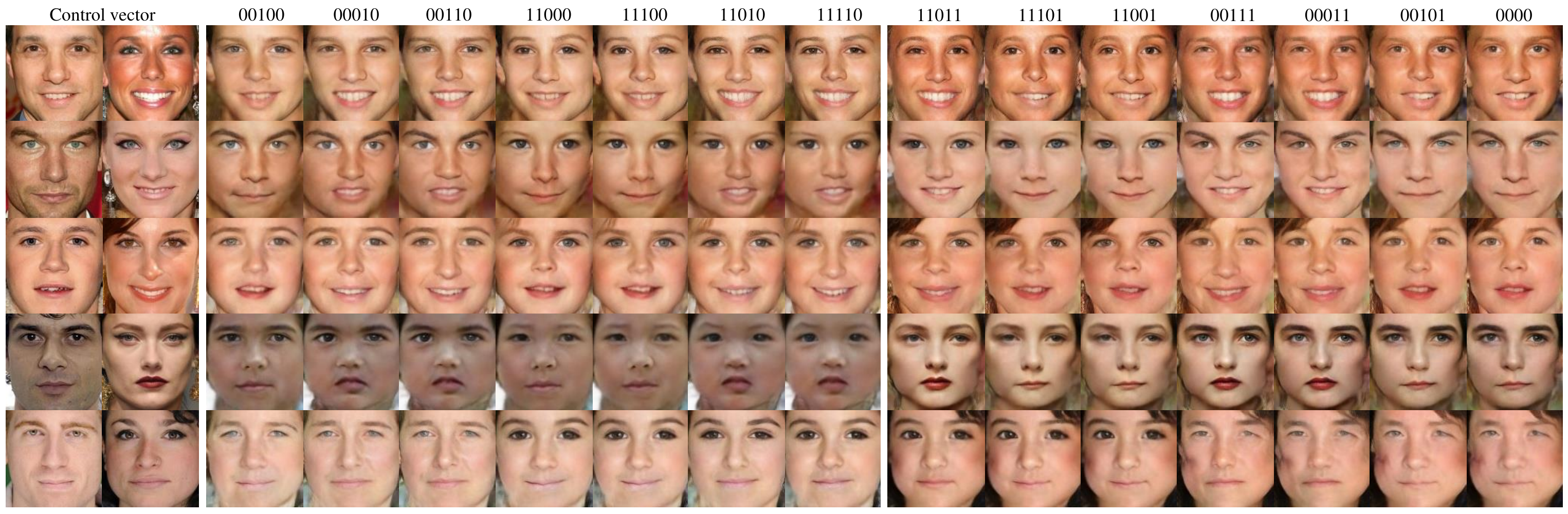}
    \scriptsize \hspace{1mm} (a) Parent faces \hspace{60mm} (b) Synthesized descendant faces \hspace{55mm}

    \caption{Synthesized descendant faces given the same parent faces under different control vectors. Parent faces in the last row come from SiblingDB and the others come from CelebAHQ. 
    Gender of descendant faces is male in the first three rows. Others are female. 
    } 
    \label{fig:inheritance}
    \vspace{-5mm}
\end{figure*}

\begin{figure*}
    \centering
\subfloat{\includegraphics[width=0.33\linewidth]{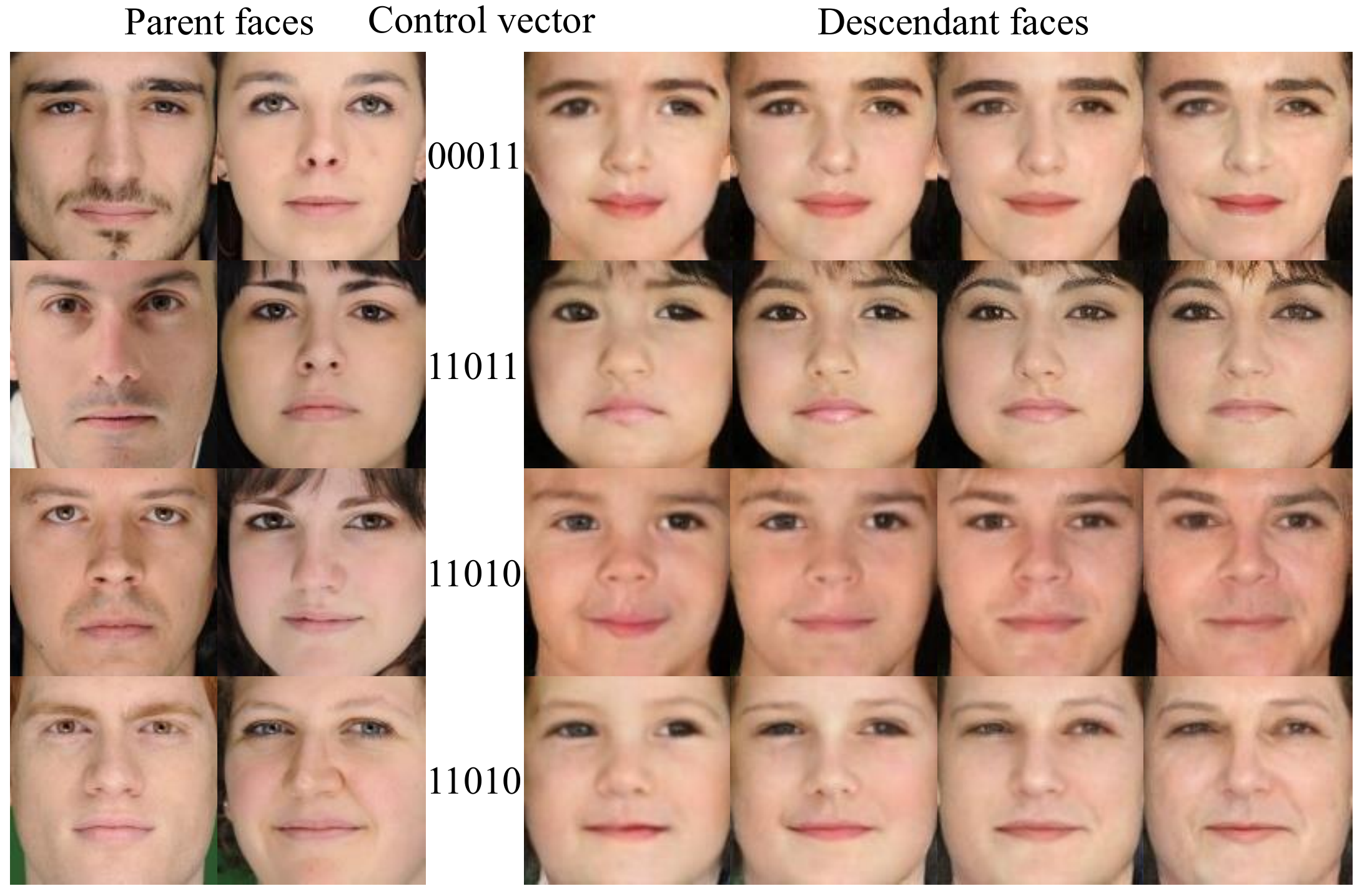}} 
\hspace{0mm}
\subfloat{\includegraphics[width=0.33\linewidth]{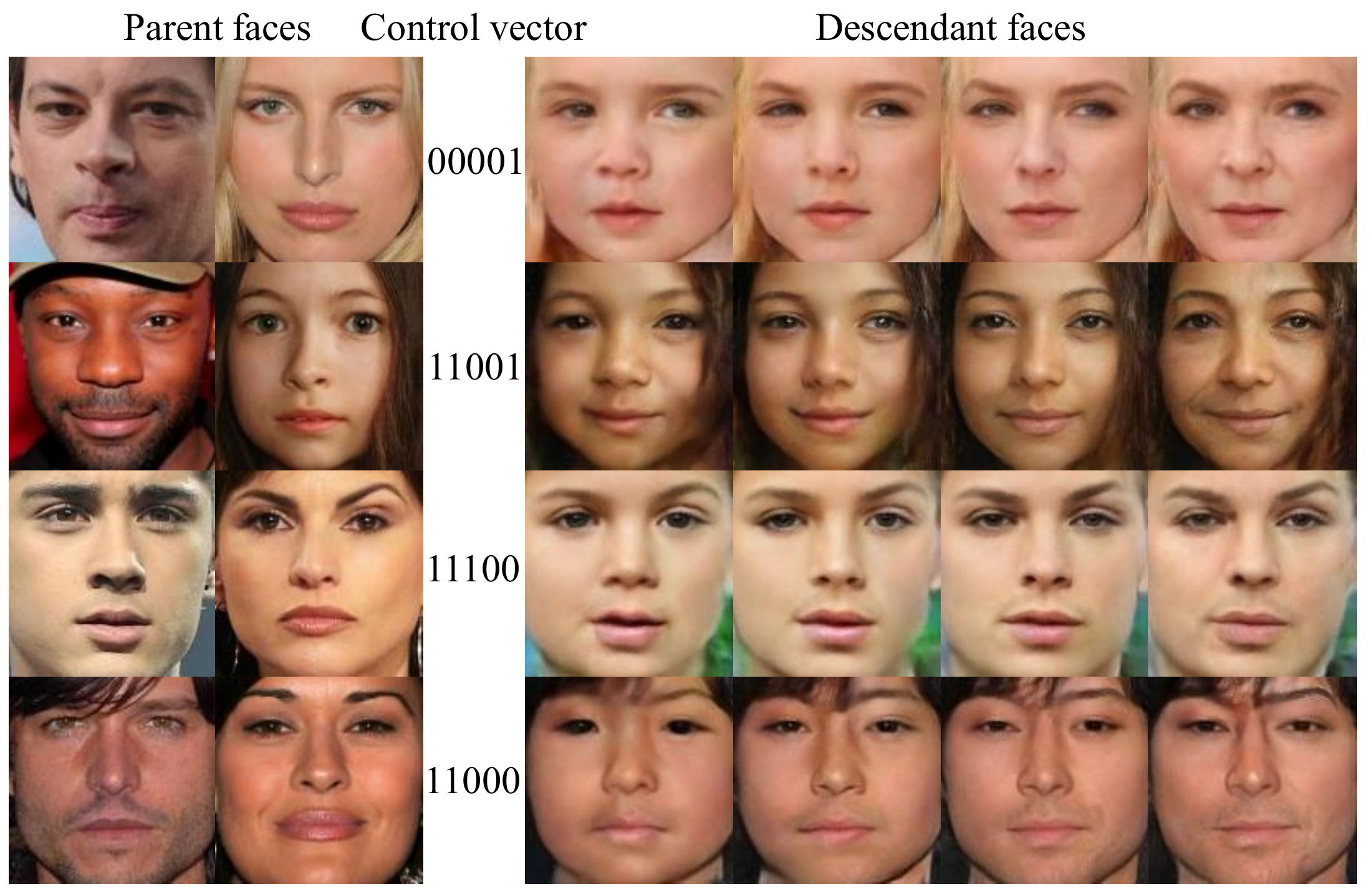}}
\hspace{0mm}
\subfloat{\includegraphics[width=0.33\linewidth]{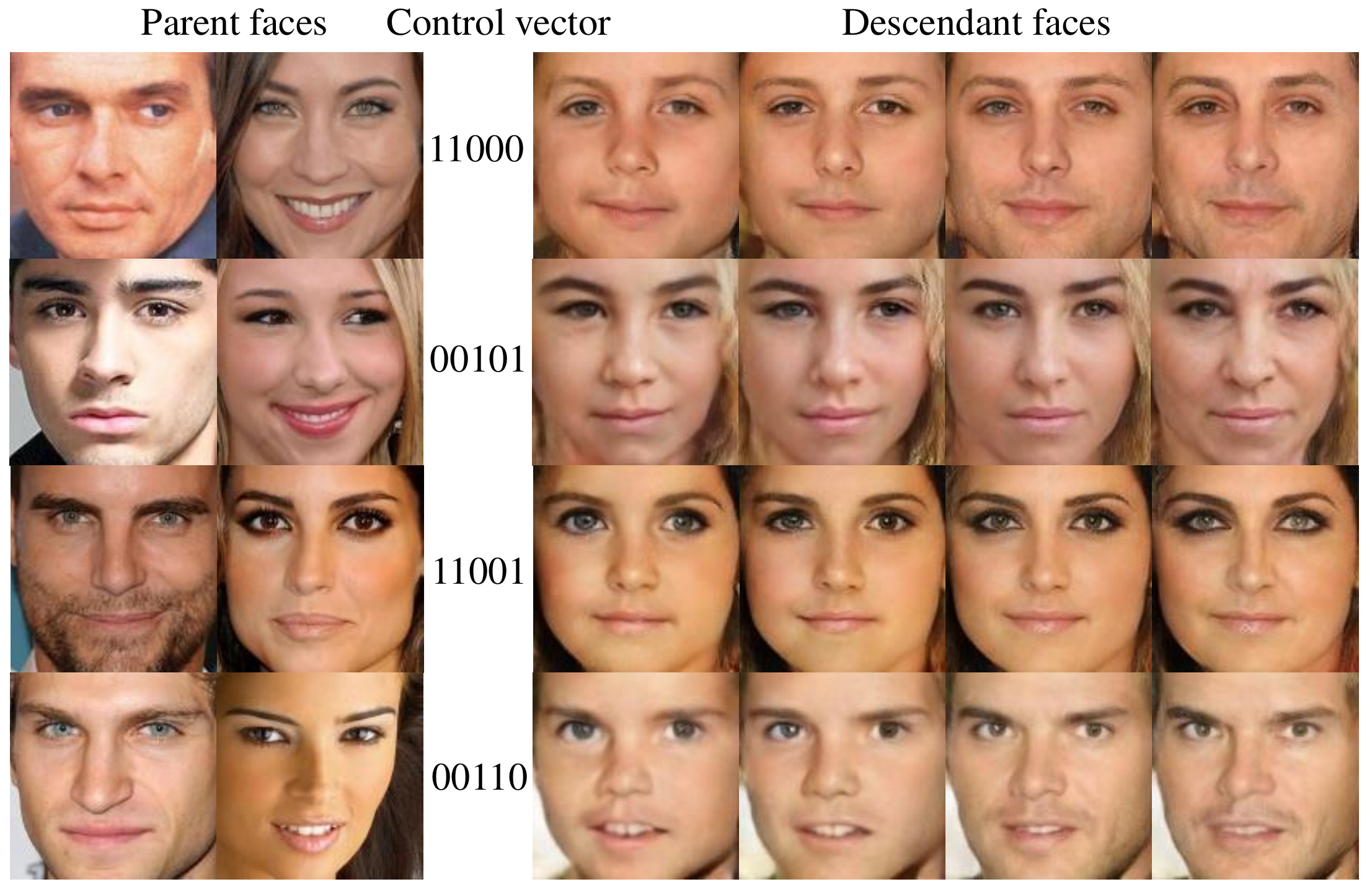}}
\vspace{0mm}
\caption{Synthesized descendant faces at different age stages.  Left: results on SiblingDB. Middle/Right: resutls on CelebAHQ. From left to right are age stage `A', `B', `C', and `D', respectively. 
	Zoom in for better view on details. }
	\label{fig:age}
	\vspace{-5mm}
\end{figure*}

\vspace{1mm}
\noindent \textbf{Preprocessing.} Facial landmarks, age and gender are provided in SiblingDB. 
CelebAHQ provides only the label of gender. 
We exploit~\cite{kazemi2014one} to detect $68$  landmarks and DEX~\cite{rothe2015dex} for age estimation. 
Faces in two databases are aligned according to the positions of two eye centers, and then cropped and resized into the size of $256 \times 256$. 
After face alignment, the positions and sizes of bounding boxes of facial components are determined. 
The sizes are $80\times96$, $80\times96$, $80\times80$, $64\times128$, and $256\times256$ for left eye\&brow, right eye\&brow, nose, mouth, and face profile, respectively. 
The inputs of our network are a pair of parent faces, a control vector of inheritance and the age and gender labels of parent faces. 
An image pair consists of a female face and a male face, which is randomly generated in two databases. 
We randomly generate 76,800 female-male face pairs and control vectors in SiblingDB, and about 4M (millions) image pairs and control vectors in CeleAHQ.

\vspace{1mm}
\noindent \textbf{Structure.} The decoder and encoder of the inheritance module have $3$ residual blocks~\cite{he2016deep}. The encoder and decoder of the attribute enhancement module have $5$ convolution layers and $1$ fully connected layer. 
Each convolution layer is followed by a max-pooling layer. 
The details about the networks are presented in the supplementary.  

\vspace{1mm}
\noindent \textbf{Training.} 
We jointly learn both modules of DFS-GAN. 
The hyperparameters are $\lambda_{0}=1$, $\lambda_{11}=10$ , $\lambda_{12}=0.1$, $\lambda_{13}=0.1$, $\lambda_{21}=0.001$, and $\lambda_{22} = 0.1$.    
We use Adam~\cite{kingma2014adam} for optimization. 
The learning rate is 0.0001 and the batch size is 8. 
The attribute enhancement module is pre-trained on UTKFace~\cite{zhang2017age} and the training sets of SiblingDB and CelebAHQ. 
UTKFace~\cite{zhang2017age} is used to compensate for the imbalanced age distribution in SiblingDB and CelebAHQ.   
We update attribute enhancement module once every 500 iterations of training inheritance module for joint learning.

\begin{figure}[t]
\centering
  \includegraphics[width=0.8\linewidth]{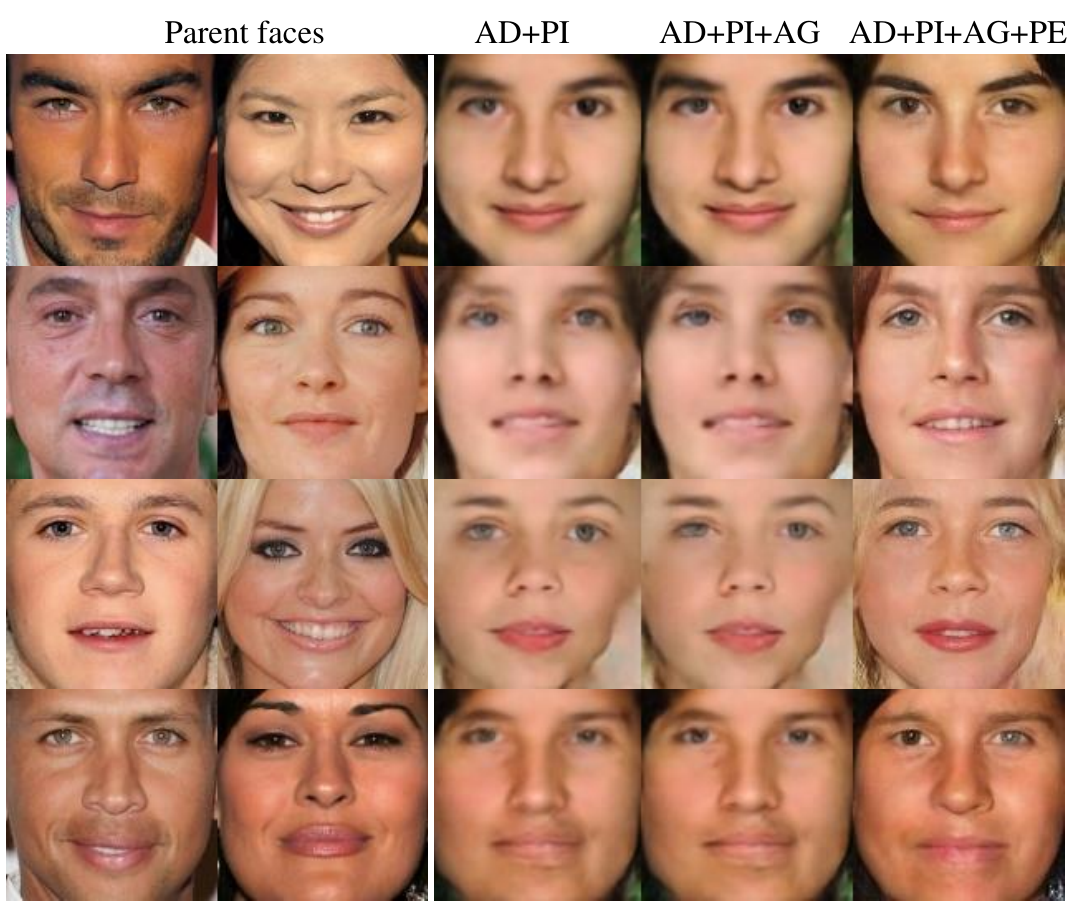}
  \caption{Ablation study of losses.}
  \label{fig:ablation}
\vspace{-10mm}
\end{figure}

\vspace{1mm}
\noindent \textbf{Ablation study.}
We have four types of losses: adversarial loss (\textbf{AD}), pixel loss (\textbf{PI}), age and gender control loss (\textbf{AG}), and perceptual loss (\textbf{PE}).
Results of using different losses are shown in Fig.~\ref{fig:ablation}, including \textbf{AD}+\textbf{PI}, \textbf{AD}+\textbf{PI}+\textbf{AG}, and \textbf{AD}+\textbf{PI}+\textbf{AG}+\textbf{PE}. 
\textbf{AD}+\textbf{PI} is the baseline. \textbf{AG} and \textbf{PE} are used for further enhancement. 
The performance gets better in terms of image quality and facial details when adding \textbf{AG} and \textbf{PE} gradually. \textbf{PE} contributes more than \textbf{AG}. 

\begin{figure*}
    \centering
\subfloat[]{\includegraphics[height=0.2\textheight]{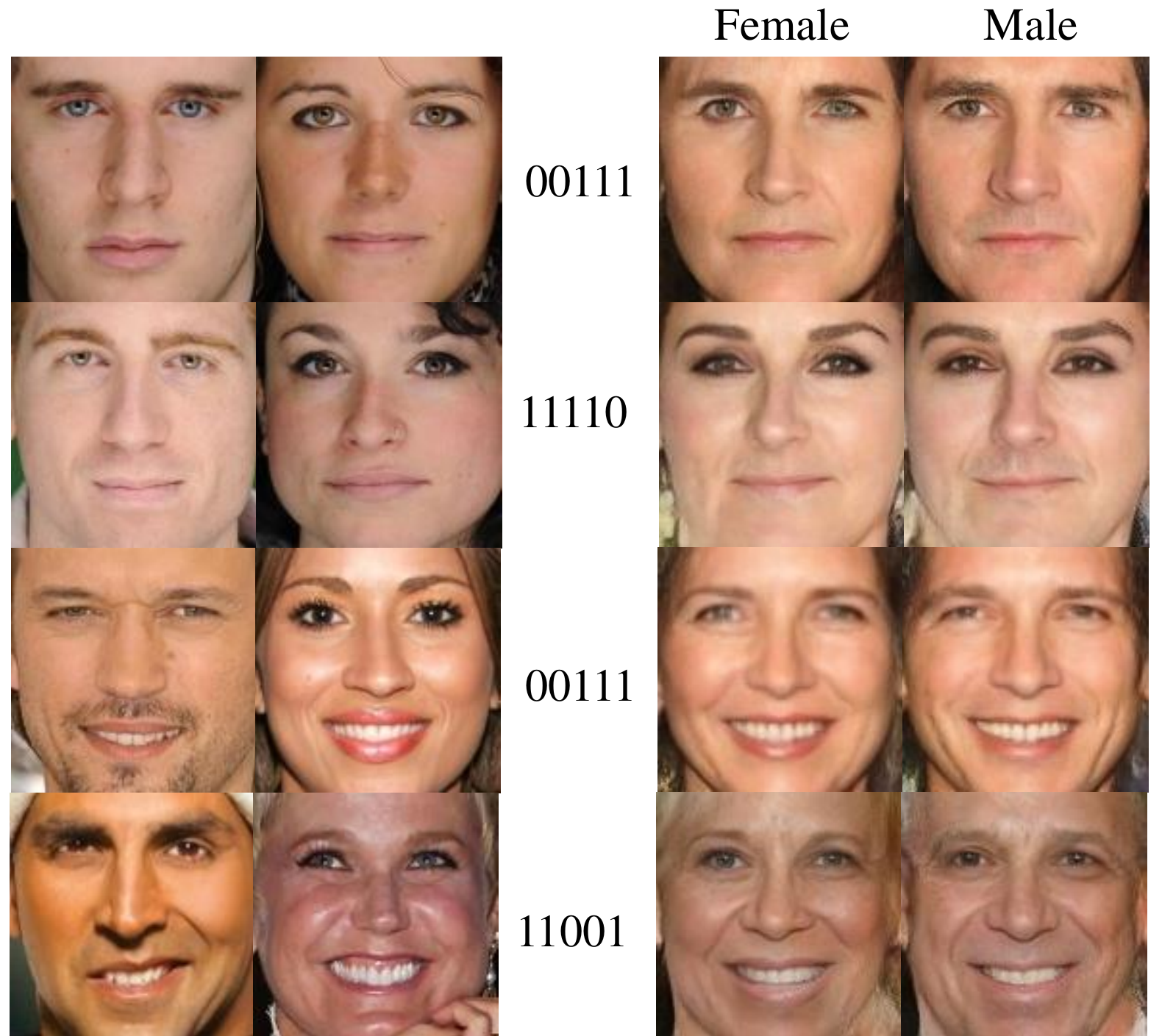}\label{fig:gender}} 
\hspace{2mm}
\subfloat[]{\includegraphics[height=0.19\textheight]{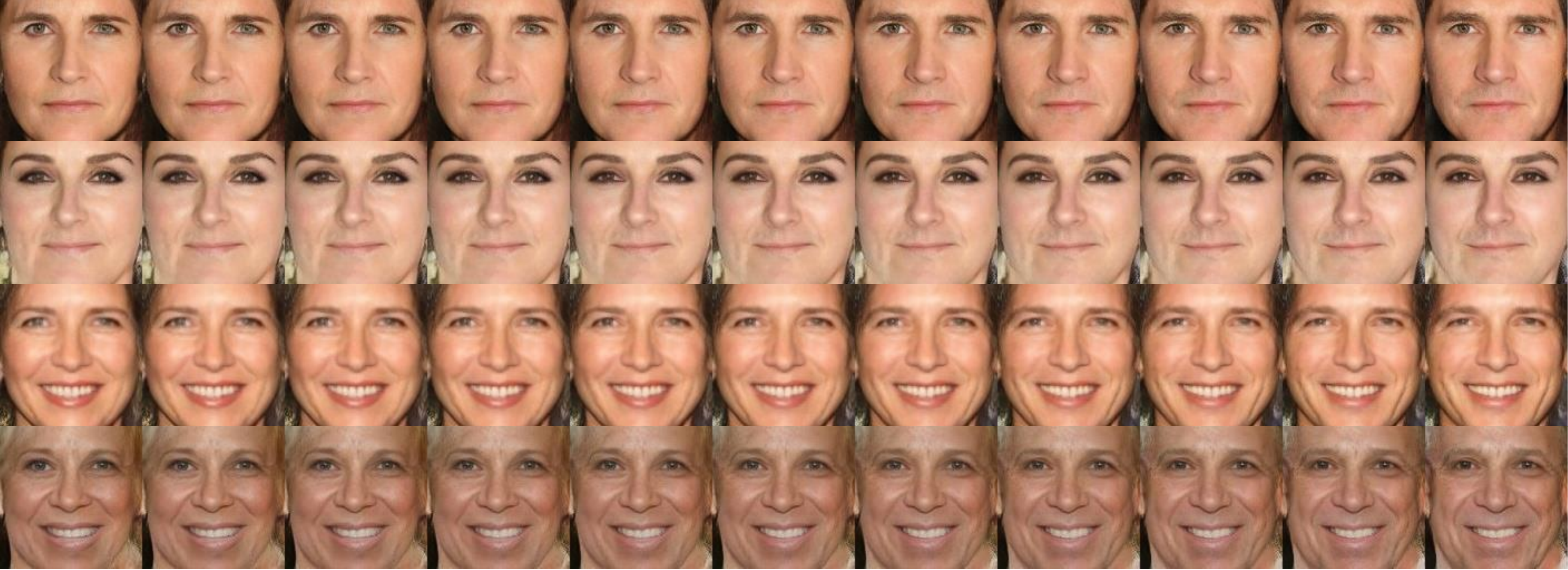}\label{fig:gender_evolution}}

\caption{(a) Synthesized descendant faces with different genders at the age stage of 'D'.
	(b) Evolution of gender from female to male.  
	Faces in top two rows are from SiblingDB. Others are from CelebAHQ. }
\vspace{-5mm}
\end{figure*}

\subsection{Visual results}
\vspace{-1mm}
\subsubsection{Control over the inheritance of components}
\vspace{-1mm}
The control vector of inheritance is used to determine the resemblance of facial components between the descendant face and its parent faces. 
As shown in Figure~\ref{fig:inheritance}, given the same parent faces, the generated descendant faces under different control vectors by our method are illustrated. 
Analyses are summarized as follows. 
Firstly, the combination of facial components is exactly according to the specified control vector. 
The descendant faces preserve the similarity of components to the corresponding components of their parent faces. 
For example, the vector `00110' means that the left and right eye\&brows inherit from the male, the nose and mouth inherit from the female, and the profile inherits from the male. 
Comparing each component of the synthesized face and parent faces, the shape and texture of eye\&brow and  profile retain the similarity to the father, while nose and mouth retain the similarity to the mother. 
Secondly, a descendant face under a control vector can be distinguished from the descendant face under another control vector according to their facial appearance. 
Thirdly, though texture, shape, color, and lighting of two parent faces are very different, our method could make the fusion of inherited components harmonically on descendant faces. 
The above analyses show that our method has accurate control over the inheritance of facial components and can generate harmonic descendant faces with retaining appearance details. 

\vspace{-2mm}
\subsubsection{Control over the age and gender}
\vspace{-1mm}
We use the attribute enhancement module to control the age and gender of the descendant face. 
Fig.~\ref{fig:age} presents descendant faces with the specified control vector of inheritance under four different age stages. 
The results show that our model captures the distinctive features of appearance under different age stages, including the shape and size of facial components, the wrinkles, the color of lips, and the tightness and glossiness of skin. 
For example, in the first row of the left figure, the tightness and glossiness of the synthesized face decrease as the age increases and the wrinkles become more noticeable.
The eyes of one child face are larger and brighter than that of an older. Besides, the redness of lips decreases as the age increases.   

For the evaluation of control over gender, Fig.~\ref{fig:gender}  presents the synthesized faces with different genders given the specified control vector. 
Fig.~\ref{fig:gender_evolution} illustrates the evolution of gender from a female face to a male face. 
The results show that our method is able to capture the differences of facial appearance between female and male descendant faces in terms of the beard, the thickness of brow, and the texture of skin. 
For example, as shown in Fig.~\ref{fig:gender}, the beard of the male descendant face becomes much more noticeable than the female as the age increases. 
The brows of the male face are thicker than the female.
The cheek of the female face is plumper than the male face. 
The evolution of these details can be observed in Fig.~\ref{fig:gender_evolution}. 
As the evolution processes, the features of the male on the descendant face become more noticeable such as the beard, the cheek and the thickness of brow. 
The above visual results demonstrate the capability of our method on the control of gender.

\vspace{-3mm}
\subsubsection{Enrich the diversity of descendant faces}
\vspace{-2mm}
As shown in Fig.~\ref{fig:inheritance}, our method has accurate control over the inheritance of each facial component. 
However, descendant faces that differ in only one component look similar when other components are nearly the same. 
Fortunately, our model is flexible to enrich the diversity of facial appearance of descendant faces by introducing random noise in the latent feature space.  
During the phase of component exchange in the inheritance module, we can select one or multiple facial components and add random noises to their latent features to increase the diversity of facial appearance. 

\begin{figure}
    \centering
    \includegraphics[width=1\linewidth]{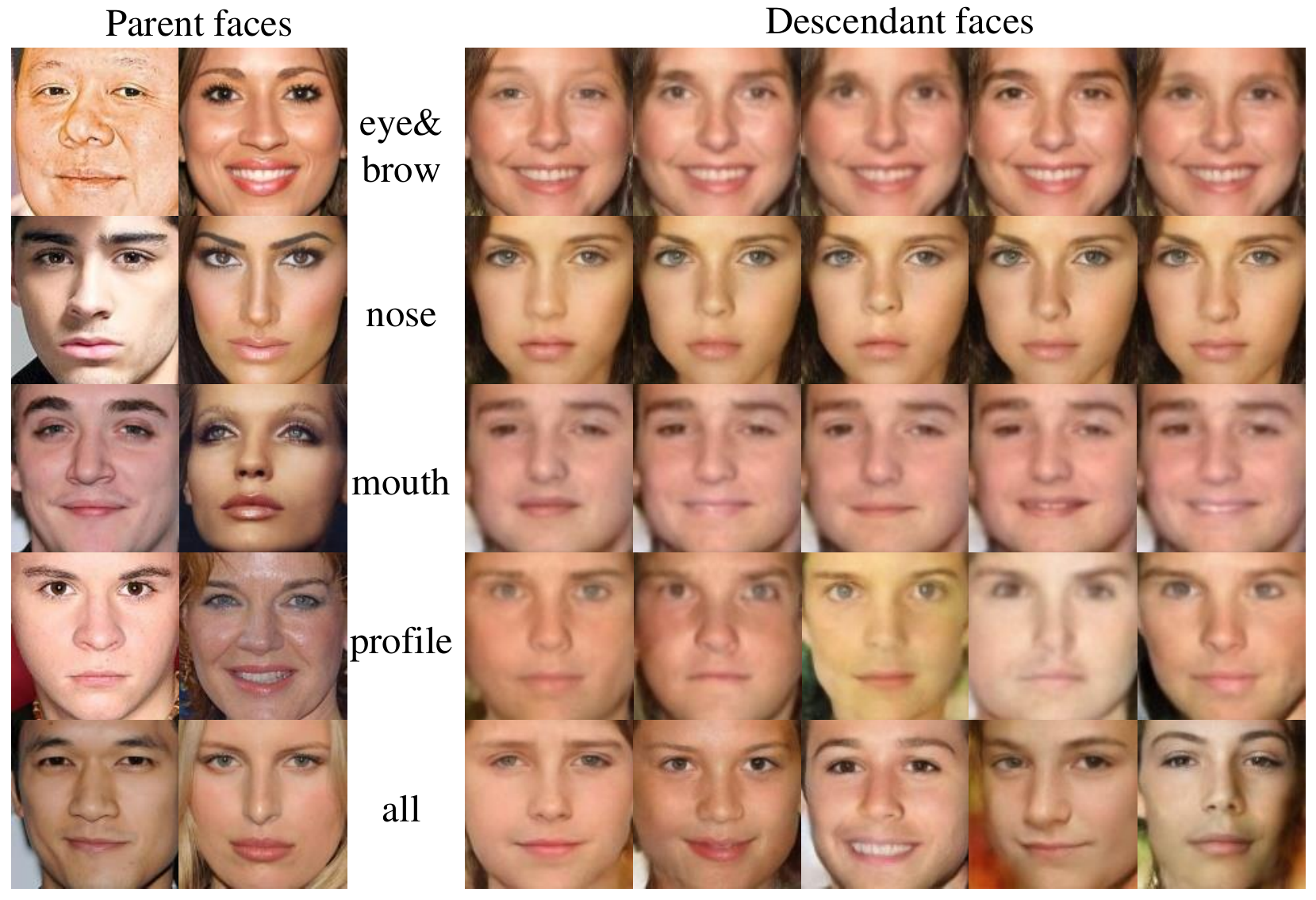}
    \caption{Enrich the diversity of descendant faces 
    } 
    \label{fig:diversity}
    \vspace{-5mm}
\end{figure}

Fig.~\ref{fig:diversity} shows the results of adding different noises on the individual component as well as all components. 
When different noises are added to an individual component, the change of its appearance is noticeable.  
For example, as shown in the last row of Fig.~\ref{fig:diversity}, when we add different noises to all components, descendant faces have a different facial appearance. 
The results show that we can increase the diversity of descendant faces by simply using random noise in our model during inference. 

\begin{table*}
	\begin{minipage}{0.69\textwidth}
	\centering
	\subfloat[Parents]{\includegraphics[height=0.52\linewidth]{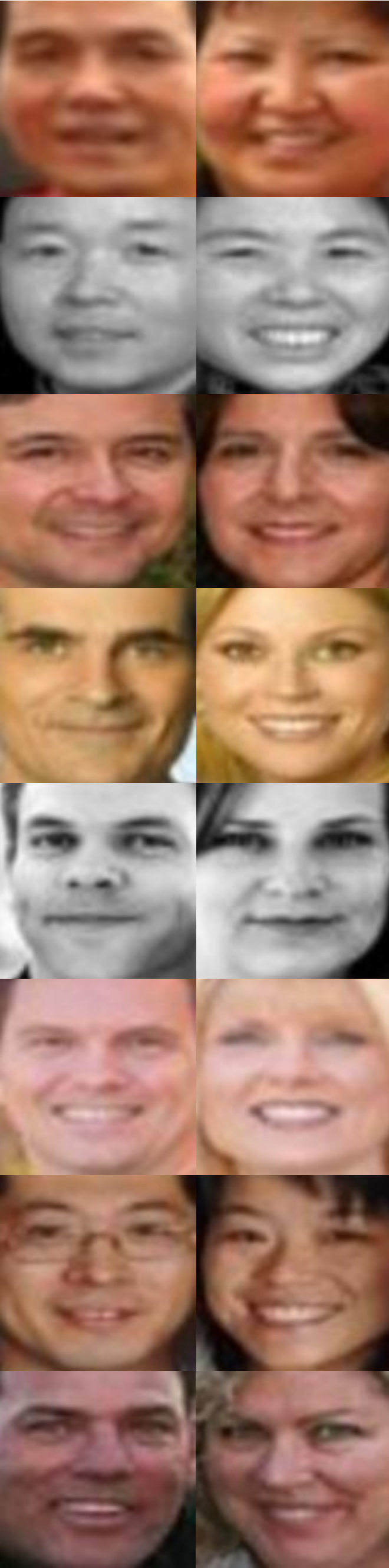}} 
	\hspace{1mm}
	\subfloat[GT]{\includegraphics[height=0.52\linewidth]{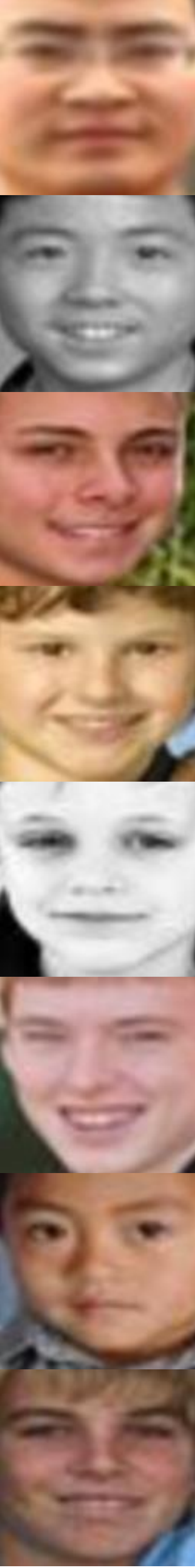}} 
	\hspace{0.1mm}
	\subfloat[~\cite{ertuugrul2018modeling}]{\includegraphics[height=0.52\linewidth]{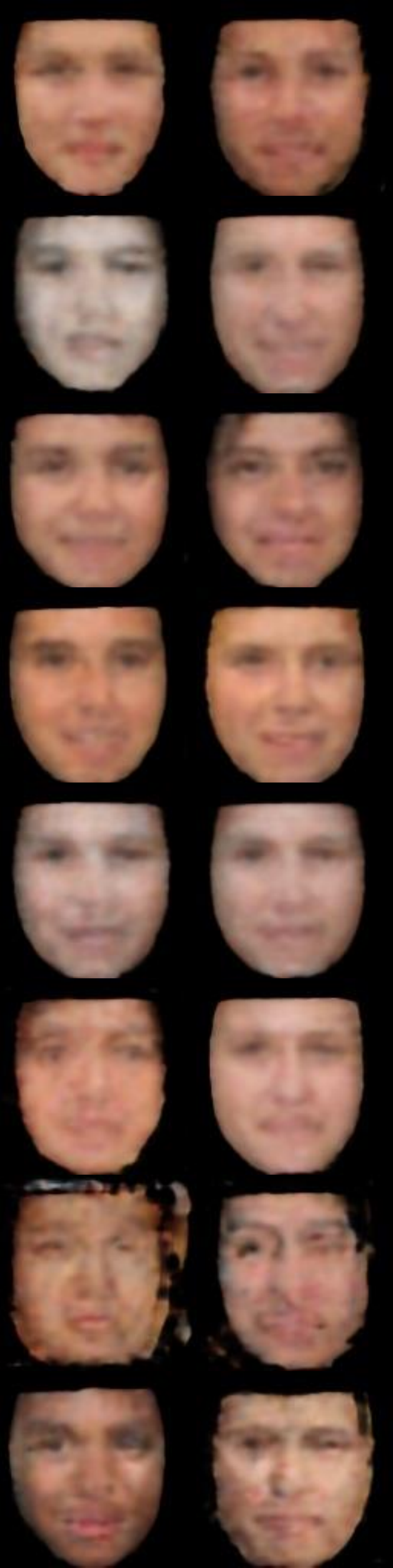}} 
	\subfloat[~\cite{ozkan2018kinshipgan}]{\includegraphics[height=0.52\linewidth]{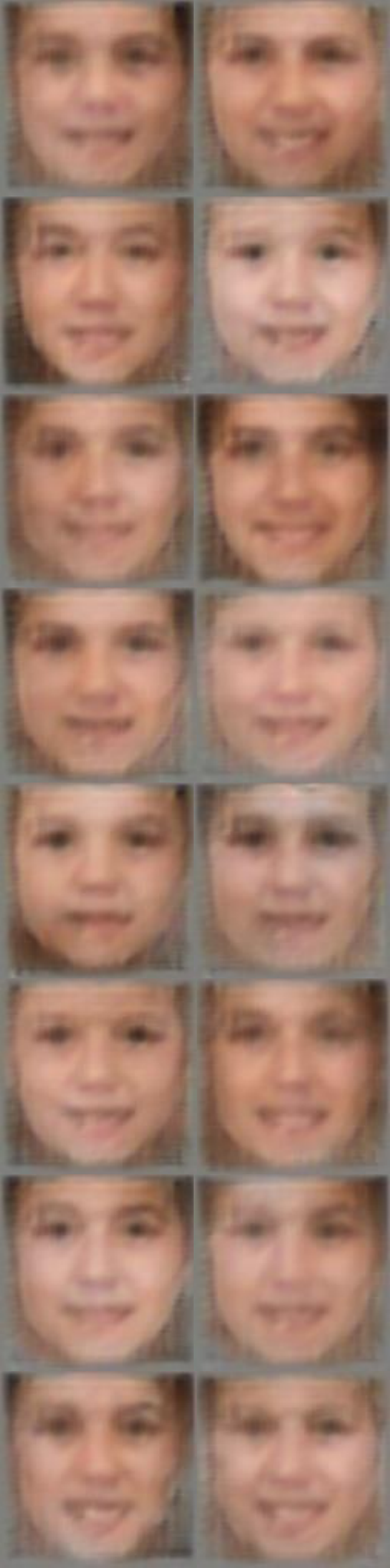}} 
	\hspace{0.1mm}
	\subfloat[Ours]{\includegraphics[height=0.52\linewidth]{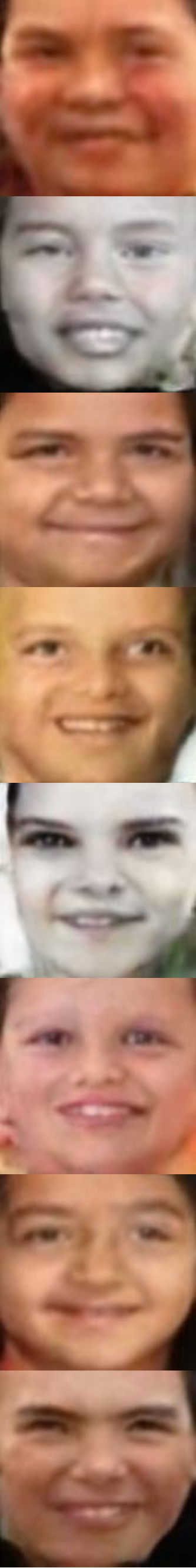}} 
	\hspace{2mm}
	\subfloat[GT]{\includegraphics[height=0.52\linewidth]{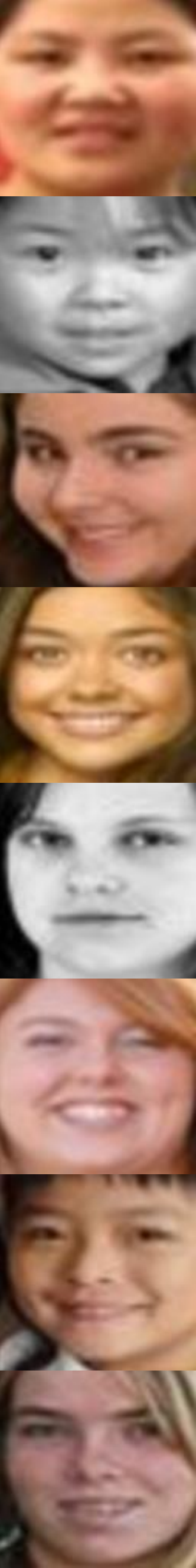}} 
	\hspace{0.1mm}
	\subfloat[~\cite{ertuugrul2018modeling}]{\includegraphics[height=0.52\linewidth]{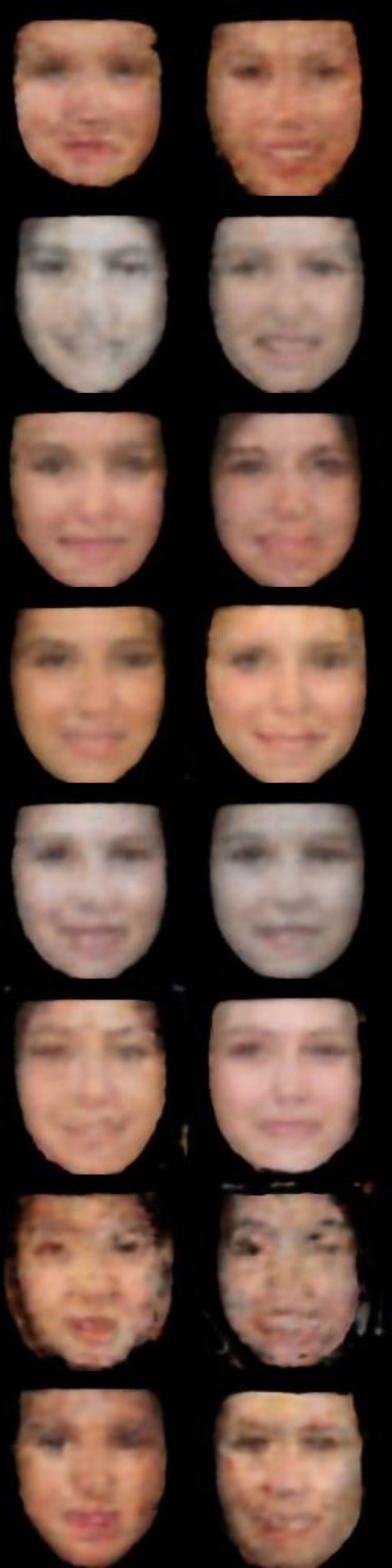}} 
	\subfloat[~\cite{ozkan2018kinshipgan}]{\includegraphics[height=0.52\linewidth]{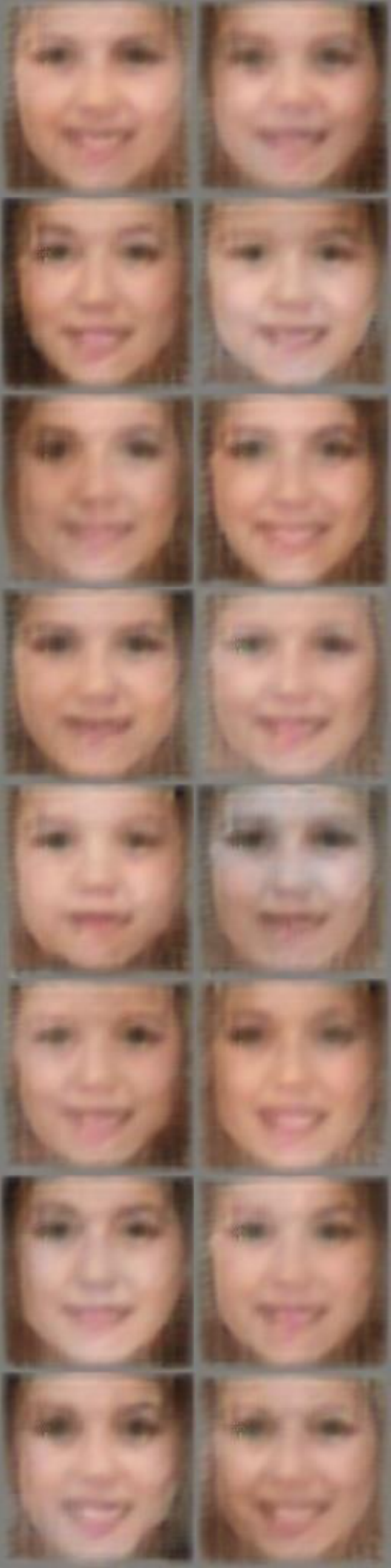}}
	\hspace{0.1mm}
	\subfloat[Ours]{\includegraphics[height=0.52\linewidth]{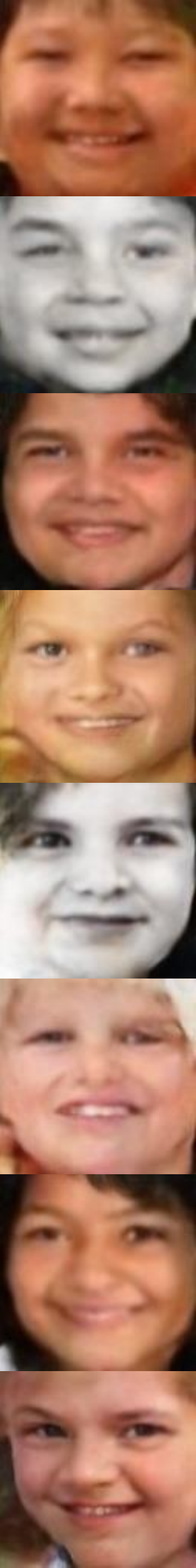}} 
	\captionof{figure}{Comparison with the state-of-the-art. (a) shows parent faces. 
		(b) and (f) are the ground truth faces of the son and daughter, respectively. 
		(c) and (g) are the results of F-S, M-S, F-D, and M-D by KPFE~\cite{ertuugrul2018modeling}. (d) and (h) are the results of of F-S, M-S, F-D, and M-D by KinshipGAN~\cite{ozkan2018kinshipgan}. 
		(e) and (i) are the results of P-S and P-D by our method.}
	\label{fig:sota}
	\end{minipage}
	\hspace{3mm}
	\begin{minipage}{0.24\textwidth}
	\centering
	\caption{Evaluation th kinship verification}
	\vspace{1mm}
	\scalebox{0.9}{
	\begin{tabular}{C{3.2cm}|C{1.5cm}}
		\hline
		Method & Verif. acc.  \\ \hline\hline
		Ground-Truth & $63.59\%$ \\ \hline \hline
		KPFE~\cite{ertuugrul2018modeling} &  $42.97\%$ \\ 
		KinshipGAN~\cite{ozkan2018kinshipgan} &  $24.69\%$\\ 
		CDFS-GAN (ours) & $\mathbf{61.88\%}$  \\\hline
	\end{tabular}}
	\label{tb:verification}
		
	\vspace{6mm}	
	\centering
	\caption{Evaluation via face verification}
	\vspace{1mm}
	\scalebox{0.85}{
	\begin{tabular}{C{3.6cm}|C{1.5cm}}
		\hline
		Method & Verif. acc. \\
		\hline \hline
		VGG-Face~\cite{parkhi2015deep} &22.0\% \\
		Microsoft Face API~\cite{microsoft} & 17.0\% \\
		Amazon ReKognition~\cite{amazon} & 9.0\% \\
		\hline 
	\end{tabular}
	}
	\label{tb:facereco}

	\vspace{10mm}

	\end{minipage}
\vspace{-4mm}
\end{table*}

\subsection{Comparison with the state-of-the-art}

The comparison with~\cite{ertuugrul2018modeling} and~\cite{ozkan2018kinshipgan} on the TSKinFace database ~\cite{qin2015tri} is shown in Fig.~\ref{fig:sota}.
Since~\cite{ertuugrul2018modeling} and~\cite{ozkan2018kinshipgan} generate a descendant face given only one parent face, we present their results of father-son (F-S), father-daughter (F-D), mother-son (M-S), and mother-daughter (M-D). 
As our model generates a descendant face given a pair of parent faces, we present the results of parents-son (P-S) and parents-daughter (P-D). 

The analyses are summarized as follows. 
Firstly, our method achieves much better image quality than competitive methods since we use the carefully designed modules while they simply exploit an auto-encoder or GAN. 
They encounter the issue that one input corresponds to multiple outputs during training. 
It messes up the network. 
We use the control vector of inheritance to alleviate this issue. 
Secondly, descendant faces generated by our method have higher similarity to the ground truth descendant faces.
Our method also keeps a better resemblance between the generated face and its parent faces. 
Thirdly, our method has better diversity in terms of the profile and the appearance of facial components. 
The profiles of synthesized faces by~\cite{ozkan2018kinshipgan} are almost the same given different input faces.

\subsection{Quantitative evaluation}\label{sec:quantative}

\textbf{Kinship verification.} To quantitatively evaluate the proposed method, we perform a cross-database kinship verification experiment. We train the kinship classifier~\cite{BMVC2015_148} on FIW~\cite{robinson2016families} and test on TSKinFace~\cite{qin2015tri}. 
The two databases have no overlap.
We apply our method and competitive methods~\cite{ertuugrul2018modeling,ozkan2018kinshipgan} on TSKinFace~\cite{qin2015tri} to synthesize descendant faces and then generate F-S, F-D, M-S, and M-D pairs for testing. 
The verification results are shown in Table~\ref{tb:verification}. 
Our method achieves much better verification accuracy than other methods. Our result is slightly worse than using the ground truth children faces for verification.
The results further demonstrate the effectiveness of the proposed method.

\textbf{Face verification.} To verify whether synthesized faces can be distinguished from parent faces, we use off-the-shelf face recognition models to perform face verification on TSKinFace~\cite{qin2015tri}, including VGG-Face~\cite{parkhi2015deep}, Microsoft Face API~\cite{microsoft}, and Amazon ReKognition API~\cite{amazon}. Verification results are shown in Table~\ref{tb:facereco}. The accuracies of the three models are low, which shows that most of the synthetic descendant faces can be distinguished from their parent faces.

\textbf{User study evaluation.} To further demonstrate the effectiveness of the proposed method, we perform three user studies to evaluate our method in terms of the resemblance of facial components, age estimation, and gender recognition.
For the resemblance, the average accuracy of identifying which parent face each component of the descendant face comes from is $97.2\%$.  
The accuracy of ranking ages of descendant faces at four stages is $91.6\%$. 
The accuracy of gender recognition is $88.8\%$. 
The user studies show that our method can capture the resemblance of a descendant face to its parent faces and capture the difference of facial appearance under different ages and genders. 
Detailed settings and results are presented in the supplementary material.

\section{Conclusion}

We propose a novel method to model two-versus-one kin relation for controllable descendant face synthesis with explicit control over the resemblance between the synthesized face and its parent faces as well as control over age and gender. 
Our model contains an inheritance module for controlling the resemblance and an attribute enhancement module for controlling age and gender.  
As the databases with father-mother-child kinship annotation are relatively small, we propose an effective strategy for model learning by using low-quality synthetic faces instead. 
Evaluations including visual results and quantitative evaluations demonstrate the effectiveness of our method. 

{\small
\bibliographystyle{ieee}
\bibliography{egbib}

\begin{thebibliography}{10}\itemsep=-1pt

\bibitem{amazon}
Amazon rekognition api.
\newblock
  \url{https://azure.microsoft.com/en-au/services/cognitive-services/face/}.

\bibitem{colorcorrection}
Color balance.
\newblock \url{https://en.wikipedia.org/wiki/Color_balance}.
\newblock Wikipedia.2018-10-17.

\bibitem{microsoft}
Microsoft face api.
\newblock
  \url{https://azure.microsoft.com/en-au/services/cognitive-services/face/}.

\bibitem{alvergne2007differential}
A.~Alvergne, C.~Faurie, and M.~Raymond.
\newblock Differential facial resemblance of young children to their parents:
  who do children look like more?
\newblock {\em Evolution and Human behavior}, 28(2):135--144, 2007.

\bibitem{baldi2012autoencoders}
P.~Baldi.
\newblock Autoencoders, unsupervised learning, and deep architectures.
\newblock In {\em ICML workshop}, 2012.

\bibitem{bitouk2008face}
D.~Bitouk, N.~Kumar, S.~Dhillon, P.~Belhumeur, and S.~K. Nayar.
\newblock Face swapping: automatically replacing faces in photographs.
\newblock In {\em TOG}. ACM, 2008.

\bibitem{chang2018pairedcyclegan}
H.~Chang, J.~Lu, F.~Yu, and A.~Finkelstein.
\newblock Pairedcyclegan: Asymmetric style transfer for applying and removing
  makeup.
\newblock In {\em CVPR}, 2018.

\bibitem{Choi2018CVPR}
Y.~Choi, M.~Choi, M.~Kim, J.-W. Ha, S.~Kim, and J.~Choo.
\newblock Stargan: Unified generative adversarial networks for multi-domain
  image-to-image translation.
\newblock In {\em CVPR}, 2018.

\bibitem{dal2006kin}
M.~F. Dal~Martello and L.~T. Maloney.
\newblock Where are kin recognition signals in the human face?
\newblock {\em Journal of Vision}, 6(12):2--2, 2006.

\bibitem{debruine2009kin}
L.~M. DeBruine, F.~G. Smith, B.~C. Jones, S.~C. Roberts, M.~Petrie, and T.~D.
  Spector.
\newblock Kin recognition signals in adult faces.
\newblock {\em Vision research}, 49(1):38--43, 2009.

\bibitem{dehghan2014look}
A.~Dehghan, E.~G. Ortiz, R.~Villegas, and M.~Shah.
\newblock Who do i look like? determining parent-offspring resemblance via
  gated autoencoders.
\newblock In {\em CVPR}, 2014.

\bibitem{dibeklioglu2017like}
H.~Dibeklioglu.
\newblock Visual transformation aided contrastive learning for video-based
  kinship verification.
\newblock In {\em ICCV}, 2017.

\bibitem{dibeklioglu2013like}
H.~Dibeklioglu, A.~Ali~Salah, and T.~Gevers.
\newblock Like father, like son: Facial expression dynamics for kinship
  verification.
\newblock In {\em ICCV}, 2013.

\bibitem{ertugrul2017will}
I.~{\"O}. Ertugrul and H.~Dibeklioglu.
\newblock What will your future child look like? modeling and synthesis of
  hereditary patterns of facial dynamics.
\newblock In {\em FG}, pages 33--40, 2017.

\bibitem{ertuugrul2018modeling}
I.~{\"O}. Ertu{\u{g}}rul, L.~A. Jeni, and H.~Dibeklio{\u{g}}lu.
\newblock Modeling and synthesis of kinship patterns of facial expressions.
\newblock {\em IVC}, 2018.

\bibitem{fang2013kinship}
R.~Fang, A.~C. Gallagher, T.~Chen, and A.~Loui.
\newblock Kinship classification by modeling facial feature heredity.
\newblock In {\em ICIP}, 2013.

\bibitem{georgopoulos2018modeling}
M.~Georgopoulos, Y.~Panagakis, and M.~Pantic.
\newblock Modeling of facial aging and kinship: A survey.
\newblock {\em IVC}, 80:58--79, 2018.

\bibitem{ghahramani2014family}
M.~Ghahramani, W.-Y. Yau, and E.~K. Teoh.
\newblock Family verification based on similarity of individual family
  member’s facial segments.
\newblock {\em Machine Vision and Applications}, 2014.

\bibitem{goodfellow2014generative}
I.~Goodfellow, J.~Pouget-Abadie, M.~Mirza, B.~Xu, D.~Warde-Farley, S.~Ozair,
  A.~Courville, and Y.~Bengio.
\newblock Generative adversarial nets.
\newblock In {\em NIPS}, 2014.

\bibitem{Gulrajani2017}
I.~Gulrajani, F.~Ahmed, M.~Arjovsky, V.~Dumoulin, and A.~Courville.
\newblock Improved training of wasserstein gans.
\newblock 2017.

\bibitem{guo2014graph}
Y.~Guo, H.~Dibeklioglu, and L.~Van~der Maaten.
\newblock Graph-based kinship recognition.
\newblock In {\em ICPR}, 2014.

\bibitem{he2016deep}
K.~He, X.~Zhang, S.~Ren, and J.~Sun.
\newblock Deep residual learning for image recognition.
\newblock In {\em CVPR}, 2016.

\bibitem{huang2017beyond}
R.~Huang, S.~Zhang, T.~Li, and R.~He.
\newblock Beyond face rotation: Global and local perception gan for
  photorealistic and identity preserving frontal view synthesis.
\newblock In {\em ICCV}, 2017.

\bibitem{kaminski2013children}
G.~Kaminski, C.~Berger, C.~Jolly, and K.~Mazens.
\newblock Children’s consideration of relevant and non-relevant facial
  features in kinship detection.
\newblock {\em LAnnee psychologique}, 113(3):321--334, 2013.

\bibitem{karras2017progressive}
T.~Karras, T.~Aila, S.~Laine, and J.~Lehtinen.
\newblock Progressive growing of gans for improved quality, stability, and
  variation.
\newblock {\em ICLR}, 2018.

\bibitem{kazemi2014one}
V.~Kazemi and J.~Sullivan.
\newblock One millisecond face alignment with an ensemble of regression trees.
\newblock In {\em CVPR}, 2014.

\bibitem{kingma2014adam}
D.~P. Kingma and J.~Ba.
\newblock Adam: A method for stochastic optimization.
\newblock {\em ICLR}, 2015.

\bibitem{korshunova2016fast}
I.~Korshunova, W.~Shi, J.~Dambre, and L.~Theis.
\newblock Fast face-swap using convolutional neural networks.
\newblock {\em ICCV}, 2017.

\bibitem{lee2017unsupervised}
D.~Lee, S.~Yun, S.~Choi, H.~Yoo, M.-H. Yang, and S.~Oh.
\newblock Unsupervised holistic image generation from key local patches.
\newblock {\em ECCV}, 2018.

\bibitem{li2017kinnet}
Y.~Li, J.~Zeng, J.~Zhang, A.~Dai, M.~Kan, S.~Shan, and X.~Chen.
\newblock Kinnet: Fine-to-coarse deep metric learning for kinship verification.
\newblock In {\em Proceedings of the 2017 Workshop on Recognizing Families In
  the Wild}, pages 13--20. ACM, 2017.

\bibitem{liu2016makeup}
S.~Liu, X.~Ou, R.~Qian, W.~Wang, and X.~Cao.
\newblock Makeup like a superstar: Deep localized makeup transfer network.
\newblock {\em IJCAI}, 2016.

\bibitem{lu2017discriminative}
J.~Lu, J.~Hu, and Y.-P. Tan.
\newblock Discriminative deep metric learning for face and kinship
  verification.
\newblock {\em TIP}, 2017.

\bibitem{lu2014neighborhood}
J.~Lu, X.~Zhou, Y.-P. Tan, Y.~Shang, and J.~Zhou.
\newblock Neighborhood repulsed metric learning for kinship verification.
\newblock {\em TPAMI}, 2014.

\bibitem{ozkan2018kinshipgan}
S.~Ozkan and A.~Orkan.
\newblock Kinshipgan: Synthesizing of kinship faces from family photos by
  regularizing a deep face network.
\newblock In {\em ICIP}, 2018.

\bibitem{parkhi2015deep}
O.~M. Parkhi, A.~Vedaldi, A.~Zisserman, et~al.
\newblock Deep face recognition.
\newblock In {\em BMVC}, 2015.

\bibitem{qin2015tri}
X.~Qin, X.~Tan, and S.~Chen.
\newblock Tri-subject kinship verification: Understanding the core of a family.
\newblock {\em TMM}, 2015.

\bibitem{robinson2016families}
J.~P. Robinson, M.~Shao, Y.~Wu, and Y.~Fu.
\newblock Families in the wild (fiw): Large-scale kinship image database and
  benchmarks.
\newblock In {\em ACM MM}, 2016.

\bibitem{rothe2015dex}
R.~Rothe, R.~Timofte, and L.~Van~Gool.
\newblock Dex: Deep expectation of apparent age from a single image.
\newblock In {\em ICCV workshops}, 2015.

\bibitem{ShrivastavaPTSW16}
A.~Shrivastava, T.~Pfister, O.~Tuzel, J.~Susskind, W.~Wang, and R.~Webb.
\newblock Learning from simulated and unsupervised images through adversarial
  training.
\newblock {\em CVRR}, 2016.

\bibitem{shu2016kinship}
X.~Shu, J.~Tang, H.~Lai, Z.~Niu, and S.~Yan.
\newblock Kinship-guided age progression.
\newblock {\em Pattern Recognition}, 59:156--167, 2016.

\bibitem{simonyan2014very}
K.~Simonyan and A.~Zisserman.
\newblock Very deep convolutional networks for large-scale image recognition.
\newblock {\em CVPR}, 2014.

\bibitem{vieira2014detecting}
T.~F. Vieira, A.~Bottino, A.~Laurentini, and M.~De~Simone.
\newblock Detecting siblings in image pairs.
\newblock {\em The Visual Computer}, 2014.

\bibitem{wang2018cross}
S.~Wang, Z.~Ding, and Y.~Fu.
\newblock Cross-generation kinship verification with sparse discriminative
  metric.
\newblock {\em TPAMI}, 2018.

\bibitem{wang2014leveraging}
X.~Wang and C.~Kambhamettu.
\newblock Leveraging appearance and geometry for kinship verification.
\newblock In {\em ICIP}, 2014.

\bibitem{wang2018face}
Z.~Wang, X.~Tang, W.~Luo, and S.~Gao.
\newblock Face aging with identity-preserved conditional generative adversarial
  networks.
\newblock In {\em CVPR}, 2018.

\bibitem{xiao2018elegant}
T.~Xiao, J.~Hong, and J.~Ma.
\newblock Elegant: Exchanging latent encodings with gan for transferring
  multiple face attributes.
\newblock {\em ECCV}, 2018.

\bibitem{yan2014discriminative}
H.~Yan, J.~Lu, W.~Deng, and X.~Zhou.
\newblock Discriminative multimetric learning for kinship verification.
\newblock {\em TIFS}, 2014.

\bibitem{BMVC2015_148}
K.~Zhang, Y.~Huang, C.~Song, H.~Wu, and L.~Wang.
\newblock Kinship verification with deep convolutional neural networks.
\newblock In {\em BMVC}, 2015.

\bibitem{zhang2017age}
Z.~Zhang, Y.~Song, and H.~Qi.
\newblock Age progression/regression by conditional adversarial autoencoder.
\newblock In {\em CVPR}, 2017.

\end{thebibliography}
}

\end{document}